\documentclass[letterpaper]{article} 
\usepackage[preprint]{aaai2027}
\usepackage[hyphens]{url}  
\usepackage{graphicx} 
\usepackage{natbib}  
\usepackage{caption} 
\usepackage{amsmath}
\usepackage{amssymb}
\usepackage{booktabs}
\usepackage{multirow}
\usepackage{array}
\newcommand{\pascalfive}{PASCAL-$5^i$}
\newcommand{\cocotwenty}{COCO-$20^i$}
\newcommand{\lvisninetytwo}{LVIS-$92^i$}
\newcommand{\isaidfive}{iSAID-$5^i$}

\newcolumntype{L}[1]{>{\raggedright\arraybackslash}p{#1}}
\newcolumntype{C}[1]{>{\centering\arraybackslash}p{#1}}
\newcolumntype{V}[1]{>{\raggedright\arraybackslash}m{#1}}
\newcolumntype{M}[1]{>{\centering\arraybackslash}m{#1}}
\newif\ifAppendixInterimData
\AppendixInterimDatafalse
\newif\ifStandaloneResultsAvailable
\StandaloneResultsAvailabletrue

\newcommand{\AppendixPluginCells}{216}
\newcommand{\AppendixPluginWins}{205}

\newcommand{\AppendixPluginMeanDelta}{+8.66}

\newcommand{\PluginSettingMinWins}{33}

\newcommand{\PluginLVISWins}{36}
\newcommand{\PluginISaidCocoCells}{36}
\newcommand{\PluginISaidCocoWins}{34}
\newcommand{\PluginISaidCocoMeanDelta}{+6.23}
\newcommand{\PluginISaidDomainWins}{34}
\newcommand{\PluginISaidDomainMeanDelta}{+6.34}
\newcommand{\PluginPolypCocoCells}{36}
\newcommand{\PluginPolypCocoWins}{34}

\newcommand{\PluginPolypCocoMeanDelta}{+10.26}
\newcommand{\PluginPolypDomainWins}{33}
\newcommand{\PluginPolypDomainMeanDelta}{+10.19}

\newcommand{\PluginFSSDINOMeanDelta}{+14.05}

\newcommand{\PluginINSIDMeanDelta}{+9.33}

\newcommand{\PluginSANSAMeanDelta}{+8.79}
\newcommand{\PluginFSSSAMMeanDelta}{+2.47}

\newcommand{\PluginBoxMeanDelta}{+5.18}
\newcommand{\PluginBoxRTwoCells}{72}

\newcommand{\PluginBoxRTwoMeanDelta}{+9.77}
\newcommand{\PluginBoxRFourCells}{72}
\newcommand{\PluginBoxRFourWins}{70}
\newcommand{\PluginBoxRFourTies}{2}
\newcommand{\PluginBoxRFourMeanDelta}{+11.02}
\newcommand{\PaperResidualLambda}{1.5}

\newcommand{\StandalonePascalMeanDelta}{+2.07}
\newcommand{\StandaloneLvisMeanDelta}{+2.70}
\newcommand{\StandaloneISaidCocoMeanDelta}{+1.51}
\newcommand{\StandaloneISaidDomainMeanDelta}{+1.86}
\newcommand{\StandalonePolypCocoMeanDelta}{+0.50}
\newcommand{\StandalonePolypDomainMeanDelta}{+1.02}

\title{SADe: Sparse-Atom Support Decontamination for Few-Shot Segmentation with Weak Support Annotations}
\author{
Hang Xing\textsuperscript{\rm 1,2},
Guangjun Liu\textsuperscript{\rm 3},
Yan Xia\textsuperscript{\rm 1,2},
Xueming Ding\textsuperscript{\rm 4}\corresponding
}
\affiliations{
\textsuperscript{\rm 1} Wuxi Key Laboratory of Photovoltaic Equipment Advanced Manufacturing Technology, Wuxi, China\\
\textsuperscript{\rm 2} Wuxi Autowell Technology Co., Ltd., Wuxi, China\\
\textsuperscript{\rm 3} Wuxi Autowell Coshin Semiconductor Technology Co., Ltd., Wuxi, China\\
\textsuperscript{\rm 4} Department of Control Science and Engineering, School of Optical-Electrical and Computer Engineering, University of Shanghai for Science and Technology, Shanghai, China\\
13605638511@163.com; xuemingding@163.com
}

\begin{document}
\maketitle

\begin{abstract}
Few-shot segmentation (FSS) commonly assumes clean pixel-level support masks, yet practical support supervision often comes as boxes, scribbles, coarse masks, or pseudo-masks. These weak annotations may include texture-similar distractors and background context alongside the target, contaminating class prototypes or visual prompts before query prediction.

We introduce SADe, a predictor-agnostic support decontamination layer that estimates the reliability of selected support patches without accessing query information. Central to SADe is sparse autoencoder (SAE) atom evidence: dense similarity may respond to both the target and texture-similar context, whereas the contrast between atom activations inside and outside the weak-support region provides factor-level reliability cues. A lightweight router combines this atom evidence with dense similarity and episode statistics to predict patch reliability and generate a cleaned support mask. Trained once on synthetic weak-support episodes constructed from FSS-1000, the router is frozen for all target evaluations. The resulting mask supports standalone prediction or can be supplied to heterogeneous FSS models through their native support interfaces without altering query-side inference.

Under a matched weak-support protocol, SADe achieves the highest query mIoU in six of nine standalone prompt--shot combinations. With the same ProMi query head, it is within 0.03 mIoU of SAM3-derived masks under tight boxes and surpasses them by 11.17 and 19.49 points under box-r2 and box-r4, respectively. As a plug-in, SADe improves over raw support in 70 of 72 matched box-family comparisons across four frozen downstream models and two datasets. On point and scribble prompts, its average performance remains close to the corresponding raw-support baseline. Ablations and atom-removal controls further show that atom evidence contributes reliability information beyond dense similarity.
\end{abstract}

\section{Introduction}

\begin{figure}[!t]
\centering
\includegraphics[width=0.96\linewidth]{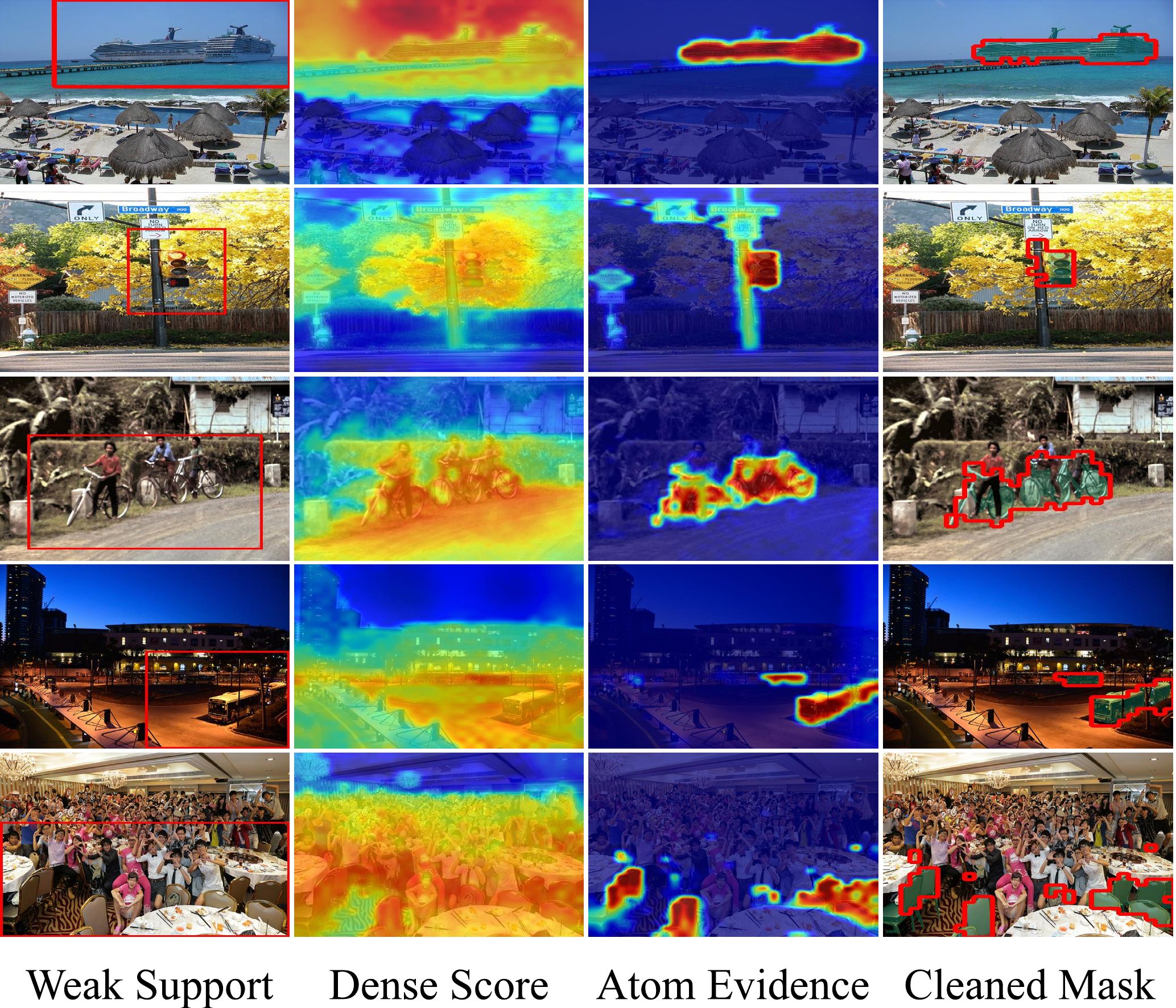}
\caption{Support-side evidence and cleaning outcomes. Each row is an episode; the final column outlines the SADe-cleaned mask in red over support GT in green to compare the cleaned region directly with the target annotation. In these examples, dense evidence responds to both target and context, whereas atom evidence is more concentrated on the target; SADe combines both to retain target patches and suppress context.}
\label{fig:mechanism_cases}
\end{figure}

\begin{figure}[t]
\centering
\includegraphics[width=\linewidth]{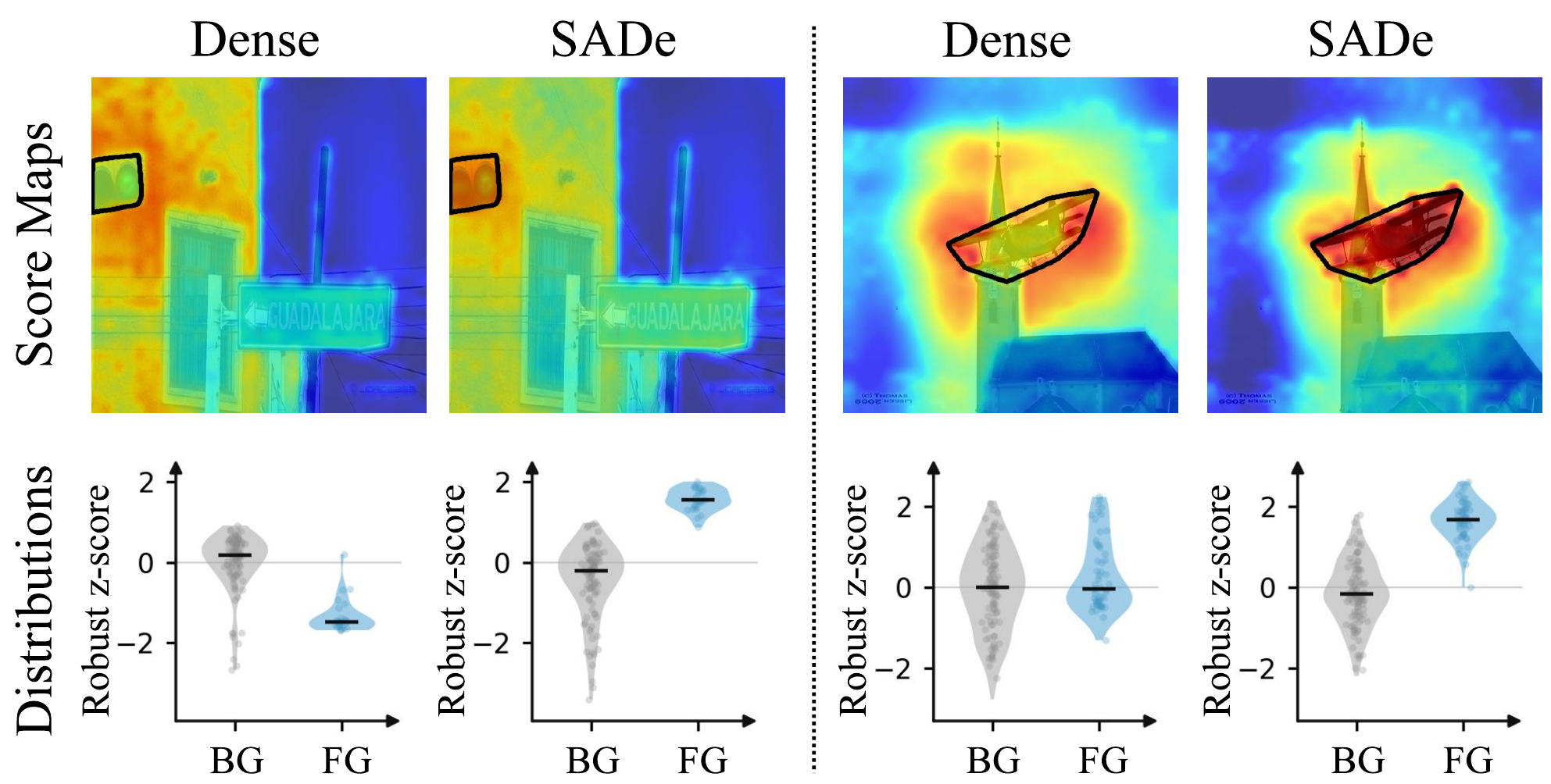}
\caption{Support-side score distributions. Violin plots show the dense prototype margin distributions before and after SADe reweighting. To place different episodes on a common scale, margins are converted to robust z-scores within each episode. Support patches are grouped into foreground and background using the GT mask, whose boundaries are shown in black. In both examples, reweighting increases FG/BG separation, indicating stronger target--context discrimination.}
\label{fig:support_distribution}
\end{figure}

Before query prediction, FSS pipelines use support annotations to guide query segmentation \citep{shaban2017oneshot}, construct prototypes \citep{wang2019panet}, provide visual prompts \citep{sun2024vrp,zhang2024gfsam,cuttano2025sansa}, or define reference sets for support-query matching \citep{min2021hsnet,liu2024matcher}. In each case, the support annotation determines which regions contribute class evidence. Weak annotations may select boundary regions, texture-similar distractors, and surrounding context together with the target, contaminating prototypes, prompts, or downstream inputs. We therefore formulate weak-support FSS as support-annotation cleaning, estimating patch reliability before selected patches become class evidence. Figure~\ref{fig:mechanism_cases} shows support-side evidence and cleaning outcomes across several episodes, while Figure~\ref{fig:support_distribution} compares Dense and SADe score maps together with the corresponding within-episode BG/FG score distributions for two examples.

\begin{figure*}
\centering
\includegraphics[width=0.98\textwidth]{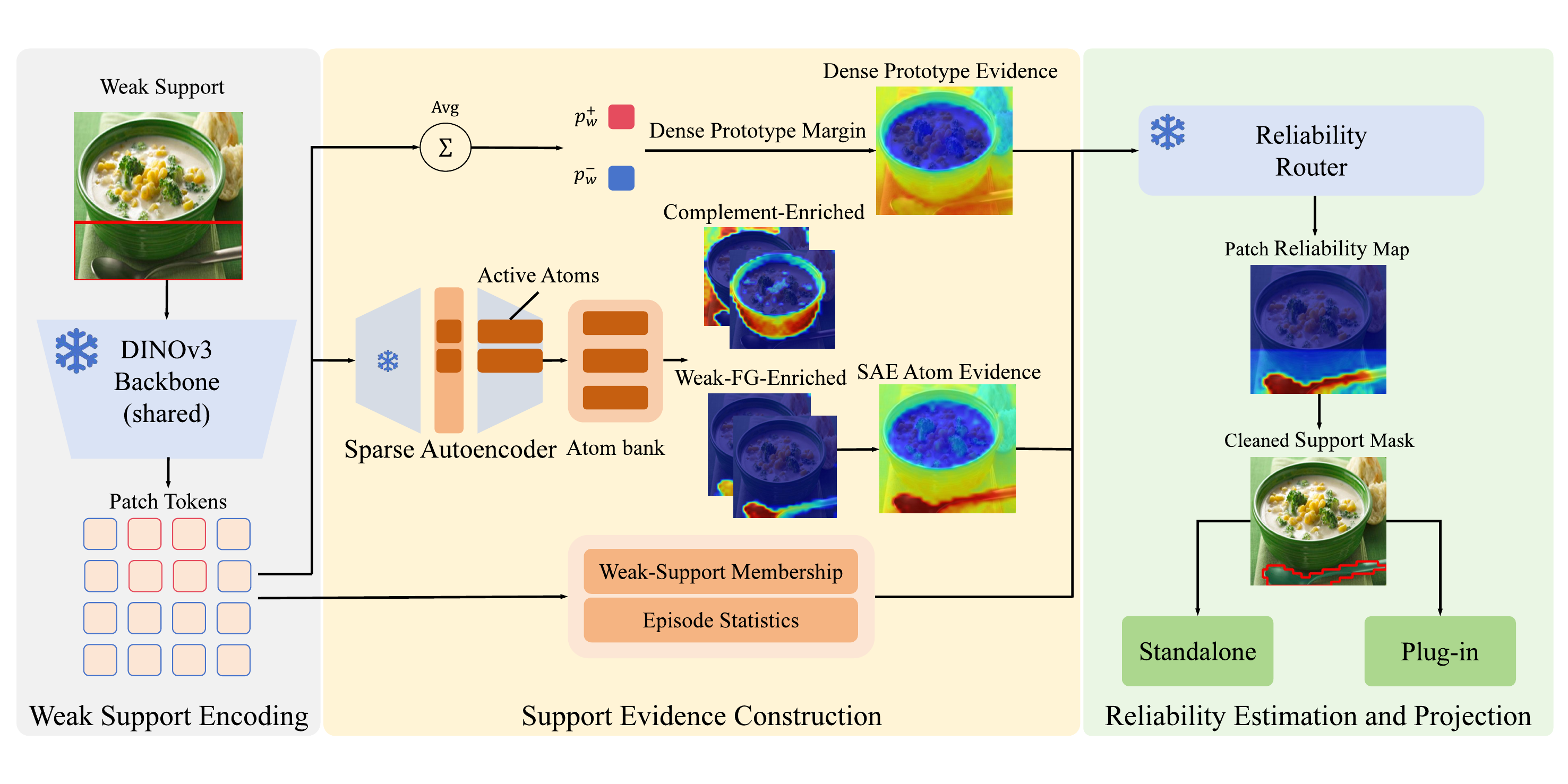}
\caption{Overview of SADe. A frozen DINOv3 backbone encodes the support image into a support-token grid, to which the weak support is mapped. SADe forms dense prototype evidence and SAE atom evidence by contrasting atom responses in the weak foreground and its complement. The reliability router fuses both evidence streams with weak-support membership and episode statistics; the resulting patch reliability map is projected into a cleaned mask for a standalone predictor or downstream FSS pipeline.}
\label{fig:overview}
\end{figure*}

Frozen self-supervised encoders provide reusable, class-agnostic patch representations for support-side reliability estimation \citep{oquab2023dinov2,simeoni2026dinov3}. On these representations, dense evidence measures similarity to the weak-support reference but may also respond to co-selected context. SADe complements it with SAE atom evidence obtained by contrasting atom activations between the weak foreground and its complement, exposing factor-level differences not captured by dense similarity \citep{lim2025patchsae,stevens2025testable,zaigrajew2025matryoshka,oh2026sparsitykey}. As summarized in Figure~\ref{fig:overview}, a router combines both signals with episode statistics to estimate patch reliability, and the resulting scores are projected into a cleaned support mask before prototype or prompt construction.

Our contributions are:
\begin{itemize}
\item We formulate weak-support FSS as support-annotation cleaning. To our knowledge, SADe is the first reusable support-annotation cleaning module for heterogeneous FSS predictors. It derives a cleaned mask from support-side information and exposes that mask through a standard interface while preserving each downstream model's native query-inference path.
\item We introduce SAE atom evidence by contrasting activations between the weak foreground and its complement to reveal factor-level cues missed by dense similarity to a noisy reference. A router combines atom and dense evidence with episode statistics to estimate patch reliability.
\item We systematically evaluate SADe under matched standalone and plug-in protocols across multiple support types, datasets, and four heterogeneous predictors. Results and ablations show broad gains under box-family prompts and confirm that atom evidence provides complementary reliability cues beyond dense similarity.
\end{itemize}
\section{Related Work}

\noindent\textbf{Few-shot segmentation under weak support.}
Few-shot segmentation uses annotated support examples to guide query segmentation or construct class representations \citep{shaban2017oneshot,wang2019panet}; later methods improve support-query matching through hypercorrelations, transductive inference, and background or base-class cues \citep{min2021hsnet,boudiaf2021repri,lang2022bam}. Recent work extends FSS support supervision to boxes and other visual prompts by unifying guidance types, conditioning promptable segmenters on visual references, or constructing prototype mixtures from boxes \citep{chang2024beyondmask,sun2024vrp,chiaroni2025promi,demarinis2025labelanything}. Others estimate or refine weak support masks: \citet{han2023boxfss} combine pseudo-trimap estimation with trimap-attention-based prototype learning; \citet{ying2026weakannotations} generate contour-based support pre-masks before prototype construction; and \citet{he2026rufnet} recalibrate support masks with query features in an integrated medical FSS model. These methods integrate weak-support handling into their support-query or predictor-specific pipelines. SADe instead decouples support cleaning from query prediction and returns a standard cleaned mask that can be reused across heterogeneous frozen predictors.

\noindent\textbf{Foundation representations for FSS.}
In FSS, foundation models can serve as frozen feature encoders, establish cross-image correspondences, or perform prompt-conditioned segmentation \citep{kirillov2023segment,ravi2024sam2,oquab2023dinov2,simeoni2026dinov3,liu2024matcher,sun2024vrp,zhang2024gfsam,cuttano2025sansa,cuttano2026insid3,park2025foreground,zakir2026fssdino,tsai2026fsssam3}. DINO-style encoders map images from different datasets into a common, class-agnostic patch feature space. SADe estimates support reliability in this shared patch space before query prediction, enabling the same cleaner to transfer across datasets and downstream models.

\noindent\textbf{Sparse autoencoders for visual representations.}
Sparse autoencoders decompose dense activations into a few active dictionary components \citep{makhzani2013ksparse}; vision studies use this factorization to expose feature structure for concept discovery, steering, intervention, and representation analysis \citep{lim2025patchsae,stevens2025testable,zaigrajew2025matryoshka,oh2026sparsitykey}. SADe applies the factorization to support-side reliability in weak-support FSS by contrasting atom activations in the weak foreground and its complement to assess whether selected patches reflect target-related or contextual factors.

\section{Method}

\subsection{Weak-Support Setting and SADe Interface}

Each episode contains support images, query images, and a target class; in multi-shot episodes, SADe independently cleans each support image while preserving the downstream method's native aggregation. SADe accepts weak support represented by a binary indicator $M_w$ on the support-token grid, including boxes, loose boxes, points, scribbles, coarse masks, and pseudo-masks. Before constructing the support sets, we exclude patches for which more than half of the pixels carry an ignore/void label. On the remaining valid patch grid, let $\mathcal{S}_w=\{i:M_w(i)=1\}$ and $\mathcal{B}_w=\{i:M_w(i)=0\}$ denote the selected patches and their complement.

Conventional weak-support prototype construction forms $p_w^+$ from $\mathcal{S}_w$ and $p_w^-$ from $\mathcal{B}_w$. If $M_w$ includes non-target patches, $p_w^+$ mixes target-relevant and contextual features; reliable class evidence must therefore be identified among selected support patches before query matching.

Let $I_s$ denote the support image, $\widetilde M_s$ the cleaned support mask, and $\theta$ the router parameters. SADe computes $\widetilde M_s$ from $I_s$ and the weak annotation $M_w$:
\begin{equation}
\widetilde M_s=\mathcal{G}_{\theta}(I_s,M_w).
\label{eq:support-cleaner}
\end{equation}
As a support-only cleaner, $\mathcal{G}_{\theta}$ is predictor-agnostic: each support input is cleaned once and its mask reused across queries. The mask feeds the shared fixed ProMi query head in standalone evaluation or is converted, when required, to the native support format of an external FSS model, while preserving its native query-inference path.

\subsection{Sparse Atom Bank and Atom Evidence}

Let $x_i$ be a patch feature extracted by a frozen vision foundation model, $\mathcal D$ a training batch, and $K$ the average number of active atoms per patch. With encoder parameters $(W_e,b_e)$, the pre-activation is $h_i=W_ex_i+b_e$. BatchTopK~\citep{bussmann2024batchtopk} retains the largest $K|\mathcal D|$ entries across the flattened batch, producing sparse codes $z_i$; decoder parameters $(W_d,b_d)$ reconstruct $\hat{x}_i$:
\begin{equation}
\{z_i\}_{i\in\mathcal D}
=\operatorname{BatchTopK}_{K}\!\left(\{h_i\}_{i\in\mathcal D}\right), \qquad
\hat{x}_i = W_d z_i + b_d.
\label{eq:sae-code}
\end{equation}
The SAE dictionary is trained on unlabeled patch features with
\begin{equation}
\mathcal L_{\mathrm{SAE}}
=\frac{1}{|\mathcal D|}\sum_{i\in\mathcal D}\|x_i-\hat{x}_i\|_2^2
+\lambda_{\mathrm{aux}}\mathcal L_{\mathrm{aux}},
\label{eq:sae-objective}
\end{equation}
where $\lambda_{\mathrm{aux}}$ weights an auxiliary loss that uses otherwise inactive atoms to reconstruct the residual and reduce dead features. For atom caching, per-patch TopK retains the $K$ atoms with the largest pre-activations in each patch. For each SAE, atoms active in at least 80\% of patches sampled from its unlabeled training pool are excluded when constructing atom evidence; the dictionary and resulting exclusion list remain fixed during target evaluation.

Let $A_i$ denote patch $i$'s retained atoms, $z_{ia}$ the activation of atom $a$, and $\epsilon>0$ a small numerical constant. SADe then compares each atom's mean activity in the weak foreground and its complement:
\begin{equation}
\mu_a^+ = |\mathcal{S}_w|^{-1}\sum_{i\in\mathcal{S}_w}z_{ia},
\qquad
\mu_a^- = |\mathcal{B}_w|^{-1}\sum_{i\in\mathcal{B}_w}z_{ia}.
\label{eq:atom-means}
\end{equation}
We summarize the weak-foreground/complement contrast with a signed atom evidence weight:
\begin{equation}
\gamma_a =
\frac{(\mu_a^+-\mu_a^-)|\mu_a^+-\mu_a^-|}
{\epsilon+\mu_a^++\mu_a^-}.
\label{eq:atom-contrast}
\end{equation}
The sign of $\gamma_a$ indicates whether atom $a$ is more active in the weak foreground or its complement, while $|\gamma_a|$ measures the normalized contrast; atoms weak or similarly active on both sides thus receive little weight. Let $T_{K_a}$ contain the $K_a$ atoms with the largest $|\mu_a^+-\mu_a^-|$, where $K_a$ is the number of episode-salient atoms retained and differs from the SAE sparsity parameter $K$. Weighting these atoms by $\gamma_a$ defines the components $u_a$ of a support-side atom evidence vector $u$ and its normalized form $\bar u$:
\begin{equation}
u_a = \gamma_a\,\mathbf{1}[a\in T_{K_a}],
\qquad
\bar u = \frac{u}{\epsilon+\|u\|_2}.
\label{eq:atom-direction}
\end{equation}
Projecting each patch's sparse activations onto $\bar u$ yields its raw atom score:
\begin{equation}
s_i^a =
\frac{\sum_{a\in A_i}\bar u_a z_{ia}}
{\epsilon+\sqrt{\sum_{a\in A_i}z_{ia}^{\,2}}}.
\label{eq:atom-score}
\end{equation}
We calibrate $s_i^a$ against the weak-complement score distribution, using its 90th percentile as the reference and its 90th-to-10th percentile range as the scale:
\begin{equation}
a_i
=
\mathcal{C}_{\mathcal{B}_w}(s_i^a)
=
\sigma\!\left(
\frac{s_i^a-Q_{0.90}(s_{\mathcal{B}_w}^a)}
{\epsilon+Q_{0.90}(s_{\mathcal{B}_w}^a)-Q_{0.10}(s_{\mathcal{B}_w}^a)}
\right).
\label{eq:atom-calibration}
\end{equation}
Here $s_{\mathcal{B}_w}^a=\{s_j^a:j\in\mathcal{B}_w\}$, $Q_p(\cdot)$ denotes the $p$-quantile, and $\sigma$ is the sigmoid; dense scores use the same weak-complement calibration. The resulting $a_i$ measures whether patch $i$ activates sparse factors more characteristic of the weak foreground than of the weak-support complement, complementing dense prototype similarity with support-side sparse-factor contrast.

\subsection{Reliability Router}

SADe next combines atom evidence $a_i$ with dense evidence to estimate the reliability of each support patch. To stabilize dense similarity, the dense branch fuses each feature with a rank-128 PCA reconstruction: $x_i^{f}=0.25x_i+0.75P_{128}(x_i)$, where $P_{128}$ denotes that reconstruction. We normalize it as $\bar x_i=x_i^{f}/(\epsilon+\|x_i^{f}\|_2)$; all dense prototypes and scores use $\bar x_i$. The atom branch is unchanged. Let $\bar p_w^+$ and $\bar p_w^-$ be the means of $\bar x_i$ over $\mathcal{S}_w$ and $\mathcal{B}_w$, respectively. The dense score and dense confidence are
\begin{equation}
s_i^d=\cos(\bar x_i,\bar p_w^+)-\cos(\bar x_i,\bar p_w^-),
\qquad
d_i=\mathcal{C}_{\mathcal{B}_w}(s_i^d).
\label{eq:dense-confidence}
\end{equation}
$s_i^d$ contrasts patch $i$'s cosine similarities to the weak-foreground and weak-complement prototypes; $\mathcal{C}_{\mathcal{B}_w}$ calibrates it against weak-complement scores to obtain $d_i$.

Support-derived atom evidence, dense evidence, and weak-support membership form the router's patch feature vector:
\begin{equation}
e_i =
[\zeta(s_i^d),\zeta(s_i^a),\pi_i^d,\pi_i^a,d_i,a_i,M_w(i)],
\label{eq:router-input}
\end{equation}
where $\zeta(\cdot)$ robustly standardizes scores within each episode, while $\pi_i^d$ and $\pi_i^a$ are the corresponding $[0,1]$-scaled percentile ranks of $s_i^d$ and $s_i^a$ over the support-token grid, encoding within-episode order and reducing sensitivity to score scale.

The episode summary $E$ provides the context needed to interpret these patch features. Its eight dimensions comprise four score means, obtained by averaging $\zeta(s_i^d)$ and $\zeta(s_i^a)$ separately over $\mathcal{S}_w$ and $\mathcal{B}_w$, together with the means and standard deviations of $d_i$ and $a_i$ within $\mathcal{S}_w$. The patch MLP then maps $(e_i,E)$ to a logit $\ell_i$, yielding $R_\theta(e_i,E)=\sigma(\ell_i)\in[0,1]$. A scalar branch maps $E$ to $\alpha(E)\in[0,1]$, which mixes $R_\theta(e_i,E)$ with $d_i$:
\begin{equation}
r_i = M_w(i)\left[\alpha(E) R_\theta(e_i,E) + (1-\alpha(E))d_i\right],
\label{eq:sade-weight}
\end{equation}
where $M_w(i)$ zeros reliability outside $\mathcal{S}_w$.

With the SAE dictionary frozen, the router is trained on FSS-1000 episodes~\citep{li2020fss1000}. For each episode, we synthesize a weak mask from the clean support mask and compute dense, atom, and weak-support membership inputs using the same evidence-construction procedure as at test time. Within $\mathcal{S}_w$, patches are labeled $y_i^s=1$ at clean-foreground coverage of at least 25\%, and $y_i^s=0$ otherwise.

The patch labels $y_i^s$ supervise local reliability; their positive fraction $\rho$ within $\mathcal{S}_w$ measures weak-support purity and provides a soft target for episode-level mixing. Lower $\rho$ raises $\alpha^\star(\rho)$, encouraging $\alpha(E)$ to shift $r_i$ toward $R_\theta(e_i,E)$; higher $\rho$ preserves more $d_i$:
\begin{equation}
\alpha^\star(\rho)
=
\operatorname{clip}_{[\alpha_{\min},\alpha_{\max}]}
\left(
\sigma\!\left(\beta_\alpha(c_\alpha-\rho)\right)
\right),
\label{eq:alpha-target}
\end{equation}
All four parameters are fixed: $c_\alpha$ sets the purity at which the target mixing weight equals $0.5$, $\beta_\alpha$ controls its decline with increasing purity, and $\alpha_{\min}$ and $\alpha_{\max}$ bound it away from $0$ and $1$.

With this episode-level target, the joint objective is
\begin{equation}
\mathcal{L}
=
\lambda_s\,\mathcal{L}_{\mathrm{wbce}}\!\left(\{\ell_i,y_i^s\}_{i\in\mathcal{S}_w}\right)
+\lambda_\alpha\operatorname{BCE}\!\left(\alpha(E),\alpha^\star(\rho)\right),
\label{eq:router-loss}
\end{equation}
where $\mathcal{L}_{\mathrm{wbce}}$ is class-balanced binary cross-entropy with logits over weak-support patches, and $\lambda_s$ and $\lambda_\alpha$ weight the patch- and episode-level terms, respectively.

\subsection{Reliability Projections}

To obtain a binary support mask, SADe sets nonfinite reliability scores to zero and $r_i^+=\max(r_i,0)$. Let $Q_p^+(r)$ denote the $p$-quantile of positive $r_i^+$ in $\mathcal{S}_w$, or zero if none exist. When $Q_{0.95}^+(r)>10^{-6}$, set
\begin{equation}
\bar r_i =
\min\!\left(1,\frac{r_i^+}{Q_{0.95}^+(r)}\right).
\label{eq:projection-normalization}
\end{equation}
Otherwise, $\bar r_i=r_i^+$; the cleaned mask retains candidates above an episode-adaptive cutoff:
\begin{equation}
\widetilde M_s(i)
=\mathbf{1}[i\in\mathcal{S}_w]\,
\mathbf{1}\!\left[
\bar r_i\geq
\max\!\left(0.5,Q_{0.35}^+(\bar r)\right)
\right].
\label{eq:hard-projection}
\end{equation}
Both protocols first apply this cutoff. In standalone evaluation, retaining fewer than three support tokens triggers a second projection without the 0.5 floor, using instead the 90th percentile of positive normalized scores. The retry is accepted only if it retains at least three support tokens and twenty valid complement tokens; otherwise, standalone reverts to the raw weak mask. In plug-in evaluation, the same retry is triggered only when the initial projection is empty; the raw weak mask is restored only if the retry also remains empty.

\section{Experiments}

\begin{table}[!t]
\centering
{\setlength{\tabcolsep}{1mm}
\begin{tabular}{@{}clrrrr@{}}
\toprule
Method & Prompt & 1-shot & 5-shot & 10-shot & Avg. \\
\midrule
 & box & 33.77 & 41.53 & 39.86 & \textbf{38.39} \\
INSID3 & box-r2 & 24.60 & 28.74 & 25.69 & 26.34 \\
 & box-r4 & 17.26 & 20.05 & 18.78 & 18.70 \\
\midrule
 & box & 23.82 & 27.58 & 25.69 & 25.70 \\
Label Anything & box-r2 & 23.34 & 27.52 & 25.51 & 25.46 \\
 & box-r4 & 22.43 & 26.81 & 24.56 & 24.60 \\
\midrule
 & box & 31.91 & \textbf{41.56} & \textbf{40.95} & 38.14 \\
ProMi & box-r2 & 22.37 & 34.99 & 36.00 & 31.12 \\
 & box-r4 & 17.44 & 27.42 & 30.43 & 25.10 \\
\midrule
\multirow{3}{*}{\shortstack{SAM3-guided\\ProMi}} & box & \textbf{33.79} & 38.67 & 37.60 & 36.69 \\
 & box-r2 & 21.09 & 27.99 & 25.26 & 24.78 \\
 & box-r4 & 12.53 & 16.22 & 14.76 & 14.50 \\
\midrule
 & box & 31.62 & 39.43 & 38.91 & 36.66 \\
SADe & box-r2 & \textbf{28.37} & \textbf{39.42} & \textbf{40.07} & \textbf{35.95} \\
 & box-r4 & \textbf{26.57} & \textbf{37.04} & \textbf{38.36} & \textbf{33.99} \\
\bottomrule
\end{tabular}
}
\caption{Standalone weak-support FSS on \cocotwenty{} under box-family prompts. Methods share folds, episodes, box annotations, validity filtering, and the evaluator while retaining their respective inference paths. Values are fold-averaged query mIoU (\%); Avg. is the 1/5/10-shot mean, and best results are bolded.}
\label{tab:standalone}
\end{table}

\subsection{Experimental Setup}

\noindent\textbf{Evaluation protocol.}
Standalone evaluation compares the end-to-end performance of complete weak-support FSS methods; within the matched ProMi subset, a shared query head isolates the effect of support-mask construction. Plug-in evaluation applies the same frozen cleaner to predictors with different architectures and native support interfaces, testing predictor-agnostic reuse. Prompts are defined on the support-token grid, where each cell corresponds to one DINOv3 patch: box tightly encloses support GT; box-r2/r4 expand it by 2/4 cells per side; point dilates a randomly sampled foreground cell by one cell; and scribble applies the same dilation to a target-crossing stroke along the foreground box's longer axis. All expansions are restricted to the valid grid. Standalone comparisons align folds, shots, support/query episodes, initial weak support, validity rules, and metrics. INSID3 retains its released inference path; Label Anything uses released fold-specific checkpoints and native box-prompt inference. SAM3-guided ProMi prompts frozen SAM3 with each support box, restricts its prediction to the box, and feeds it to the matched ProMi head. We evaluate 1/5/10-shot episodes under the standard \cocotwenty{} split~\citep{lin2014coco,nguyen2019feature}.

Plug-in evaluation pairs Raw and SADe support inputs within each episode while fixing the predictor checkpoint, feature extraction, query inference, and evaluator, so only the support input changes. FSSDINO, INSID3, and SANSA receive masks through their native interfaces; FSS-SAM3 follows its released collage-based inference and receives the tight bounding box derived from each mask. We retain the released 1/5-shot layouts and use the same fixed vertical 10-shot layout for Raw and SADe. Evaluation covers 1/5/10-shot episodes on \cocotwenty{} and \pascalfive{}~\citep{shaban2017oneshot} and reports query mIoU (\%).

\begin{table*}[!t]
\centering
{\footnotesize
\newcommand{\dashcell}{\multicolumn{1}{c}{--}}
\setlength{\tabcolsep}{1mm}
\begin{tabular}{@{}cc|rrrrrr|rrrrrr@{}}
\toprule
 & & \multicolumn{6}{c|}{\cocotwenty{}} &
\multicolumn{6}{c}{\pascalfive{}} \\
\cmidrule(lr){3-8}\cmidrule(l){9-14}
Method & Prompt & \multicolumn{2}{c}{1-shot} & \multicolumn{2}{c}{5-shot} & \multicolumn{2}{c|}{10-shot} &
\multicolumn{2}{c}{1-shot} & \multicolumn{2}{c}{5-shot} & \multicolumn{2}{c}{10-shot} \\
\cmidrule(lr){3-4}\cmidrule(lr){5-6}\cmidrule(lr){7-8}\cmidrule(lr){9-10}\cmidrule(lr){11-12}\cmidrule(l){13-14}
 & & Raw & SADe & Raw & SADe & Raw & SADe &
Raw & SADe & Raw & SADe & Raw & SADe \\
\midrule
\multirow{4}{*}{FSSDINO} & box & 30.29 & \textbf{36.92} & 37.29 & \textbf{43.63} & 35.44 & \textbf{43.22} & 56.82 & \textbf{63.57} & 61.97 & \textbf{71.18} & 63.19 & \textbf{73.88} \\
& box-r2 & 22.64 & \textbf{30.26} & 26.34 & \textbf{40.28} & 26.88 & \textbf{39.83} & 43.88 & \textbf{61.80} & 48.10 & \textbf{70.94} & 48.38 & \textbf{72.96} \\
& box-r4 & 18.42 & \textbf{27.47} & 20.95 & \textbf{36.29} & 20.77 & \textbf{36.77} & 36.05 & \textbf{57.83} & 37.64 & \textbf{66.72} & 36.55 & \textbf{68.82} \\
& sparse & 28.20 & \textbf{29.63} & 33.05 & \textbf{33.85} & \textbf{35.73} & 35.30 & \textbf{36.92} & 35.44 & \textbf{38.18} & 34.53 & \textbf{47.80} & 41.37 \\
\midrule
\multirow{4}{*}{INSID3} & box & 34.31 & \textbf{38.04} & 42.44 & \textbf{45.47} & 40.11 & \textbf{44.11} & 62.43 & \textbf{67.23} & 72.24 & \textbf{75.78} & 72.99 & \textbf{77.74} \\
& box-r2 & 25.10 & \textbf{32.10} & 29.70 & \textbf{42.68} & 27.57 & \textbf{42.02} & 46.27 & \textbf{64.81} & 52.96 & \textbf{75.76} & 50.67 & \textbf{77.22} \\
& box-r4 & 18.42 & \textbf{28.48} & 21.71 & \textbf{37.30} & 20.07 & \textbf{39.25} & 33.28 & \textbf{60.14} & 36.71 & \textbf{72.48} & 33.49 & \textbf{76.03} \\
& sparse & 26.45 & \textbf{27.21} & \textbf{26.61} & 24.22 & \textbf{25.05} & 20.95 & \textbf{38.85} & 36.82 & \textbf{28.22} & 23.33 & \textbf{25.57} & 20.79 \\
\midrule
\multirow{4}{*}{SANSA} & box & 43.40 & \textbf{48.30} & 47.34 & \textbf{54.74} & 43.87 & \textbf{52.22} & 68.61 & \textbf{71.85} & 74.09 & \textbf{77.71} & 74.54 & \textbf{79.66} \\
& box-r2 & 31.85 & \textbf{39.48} & 33.19 & \textbf{47.38} & 31.33 & \textbf{44.25} & 58.03 & \textbf{70.92} & 60.89 & \textbf{77.23} & 62.98 & \textbf{79.10} \\
& box-r4 & 26.17 & \textbf{35.85} & 26.97 & \textbf{40.91} & 26.47 & \textbf{39.05} & 49.77 & \textbf{67.80} & 50.89 & \textbf{73.87} & 52.75 & \textbf{76.50} \\
& sparse & 35.53 & \textbf{36.59} & 44.52 & \textbf{47.07} & 43.63 & \textbf{45.43} & \textbf{41.95} & 39.37 & \textbf{50.33} & 48.09 & \textbf{51.74} & 49.30 \\
\midrule
\multirow{4}{*}{FSS-SAM3} & box & 64.58 & \textbf{65.09} & 67.04 & \textbf{67.71} & 64.75 & \textbf{64.86} & 87.50 & \textbf{87.76} & \textbf{88.15} & 88.01 & \textbf{88.21} & 87.83 \\
& box-r2 & 62.78 & \textbf{63.52} & 61.42 & \textbf{64.34} & 61.57 & \textbf{63.80} & 86.57 & \textbf{86.67} & 87.39 & \textbf{88.27} & 87.31 & \textbf{88.53} \\
& box-r4 & 62.74 & \textbf{62.90} & 57.51 & \textbf{62.38} & 56.57 & \textbf{61.01} & 86.13 & \textbf{86.60} & 84.01 & \textbf{87.87} & 84.86 & \textbf{88.30} \\
& sparse & 66.25 & \textbf{66.63} & \textbf{67.01} & 66.78 & \textbf{62.81} & 62.40 & \textbf{85.28} & 85.26 & \textbf{82.55} & 81.73 & \textbf{81.14} & 80.00 \\
\bottomrule
\end{tabular}
}
\caption{Paired plug-in results on \cocotwenty{} and \pascalfive{}. Raw/SADe pairs differ only in support input. Values are fold-averaged query mIoU (\%); sparse averages point/scribble, and the better result in each pair is bolded.}
\label{tab:plugin}
\end{table*}

\noindent\textbf{Implementation details.}
All downstream FSS models remain frozen. SADe extracts layer-11 patch features with a frozen DINOv3 ViT-B/16 after direct $512\times512$ resizing without padding. In the matched standalone comparison, the fixed ProMi query head applies the same dense fusion to support and query; plug-in predictors retain their native query-inference paths. The SAE used for target-dataset evaluation has a dictionary of 16k atoms. It is trained on unlabeled COCO train2014 DINOv3 features using BatchTopK budget 32, batch size 4096, learning rate $3\times10^{-4}$, a 500-step warmup, and 8000 steps, then frozen for \cocotwenty{} and \pascalfive{}.

The 44,784-parameter router is a hidden-dimension-96 MLP trained with an 800-step budget on synthetic FSS-1000 weak-support episodes with AdamW (learning rate $3\times10^{-4}$; weight decay $10^{-4}$). The final router checkpoint is selected on held-out FSS-1000 classes and then frozen for all target evaluations. Training uses an independent 8k-atom FSS-1000 SAE (BatchTopK budget 24) and a rank-128 PCA fitted on the same features. Because the router consumes scalar atom evidence, it transfers directly to the 16k-atom COCO SAE without dictionary alignment. Target evaluation uses this SAE and a rank-128 PCA fitted on COCO train2014, while \cocotwenty{} episodes come from val2014. No \cocotwenty{} or \pascalfive{} episodes, annotations, or fold definitions are used to optimize the router weights or select its checkpoint.

\subsection{Standalone Weak-Box FSS}

Table~\ref{tab:standalone} compares SADe with INSID3~\citep{cuttano2026insid3}, Label Anything~\citep{demarinis2025labelanything}, and two matched support-mask controls. In the matched ProMi-based comparison, ProMi, SAM3-guided ProMi, and SADe use the same public prototype-mixture query rule~\citep{chiaroni2025promi} and DINOv3 feature cache; only the support mask passed to the shared head differs: the initial weak-support mask, a mask predicted by SAM3~\citep{carion2025sam3segmentconcepts}, or the SADe-cleaned mask.

SADe achieves the highest query mIoU in six of nine prompt--shot combinations, covering all box-r2 and box-r4 settings. Within the matched ProMi-based controls, average query mIoU spans only 1.48 points across the three support-mask constructions under tight boxes. Under box-r2 and box-r4, SADe's margins over SAM3-guided ProMi are 11.17 and 19.49 points, while its margins over raw-mask ProMi are 4.83 and 8.89 points. Raw-mask ProMi retains all context selected by the weak box. SAM3-guided ProMi remains competitive under tight boxes but falls below SADe on box-r2 and box-r4 as these prompts admit more context. Together with the ablations below, the widening gap is consistent with their different support-mask construction mechanisms: box expansion weakens the spatial constraint that SAM3's box prompt imposes on the target region, whereas SADe combines dense similarity with foreground/complement atom contrast to identify reliable patches and suppress target-irrelevant context.

\subsection{Plug-In Evaluation Across Predictors}

Table~\ref{tab:plugin} reports paired plug-in results for FSSDINO, INSID3, SANSA, and FSS-SAM3 while preserving each model's native support interface.

\begin{figure*}[!t]
\centering
\includegraphics[width=0.92\textwidth]{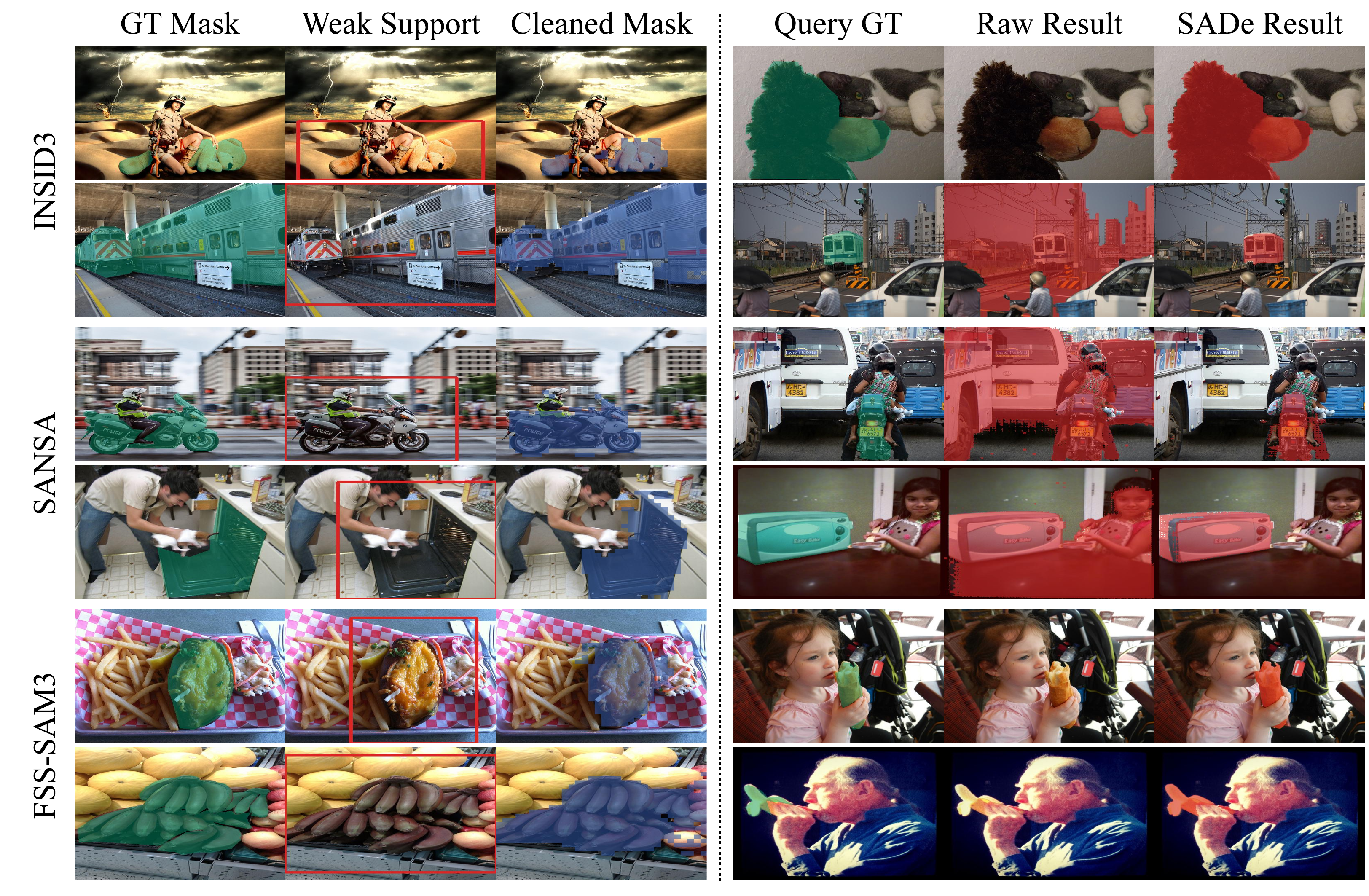}
\caption{\cocotwenty{} plug-in examples under box-r4 weak support, grouped by downstream model. Support/query GT are green, weak boxes and predictions red, and cleaned support blue. Each Raw/SADe pair shares the episode, frozen model, and query inference; only support input differs.}
\label{fig:downstream_examples}
\end{figure*}

Across four predictors, SADe improves query mIoU in 77 of 96 reported settings, including 70 of 72 box-family settings and all 48 box-r2/box-r4 settings. In every predictor--dataset combination, aggregate gains are larger under expanded boxes than under tight boxes. Because box expansion admits more target-irrelevant context into the weak support, this trend indicates that the benefit of SADe's support cleaning becomes more pronounced as such context increases. More importantly, SADe improves all 24 box-r2/box-r4 comparisons on \pascalfive{}. The router used in this evaluation is trained and selected on FSS-1000; the SAE dictionary and rank-128 PCA are trained and fitted, respectively, on COCO train2014 features. Together, these results show that SADe transfers across datasets and downstream predictors without PASCAL-specific training, fitting, or model selection. Point and scribble prompts are unseen during router training and checkpoint selection and contain substantially less surrounding context than box-family prompts; they therefore test both transfer to unseen prompt forms and preservation of downstream performance when little context contamination is present. SADe remains close to Raw overall on these sparse prompts, showing that support cleaning has limited aggregate impact when little removable context is present. Figure~\ref{fig:downstream_examples} illustrates representative paired improvements across downstream pipelines.

\subsection{Ablation and Support-Side Evidence}

\begin{table}[!t]
\centering
{\footnotesize
\setlength{\tabcolsep}{3.0pt}
\begin{tabular}{@{}ccc|ccccc@{}}
\toprule
Dense & Atom & Router & $20^0$ & $20^1$ & $20^2$ & $20^3$ & Mean \\
\midrule
$\checkmark$ &  &  & 28.47 & 35.17 & 30.88 & 30.42 & 31.24 \\
 & $\checkmark$ &  & 29.03 & 36.05 & 31.73 & 31.17 & 32.00 \\
$\checkmark$ & $\checkmark$ &  & 29.76 & 36.65 & 32.07 & 31.70 & 32.55 \\
$\checkmark$ &  & $\checkmark$ & 29.49 & 36.54 & 32.88 & 32.50 & 32.85 \\
$\checkmark$ & $\checkmark$ & $\checkmark$ & \textbf{30.37} & \textbf{38.07} & \textbf{34.07} & \textbf{33.45} & \textbf{33.99} \\
\bottomrule
\end{tabular}
}
\caption{Component ablation on \cocotwenty{} box-r4. Values are per-fold query mIoU (\%) averaged over 1/5/10-shot. Dense/Atom denote evidence inputs and Router the learned estimator; all rows share the projection rule, and best values are bolded.}
\label{tab:ablation}
\end{table}

\begin{table}[!t]
\centering
{\footnotesize
\begin{tabular}{@{}lrrr@{}}
\toprule
\multicolumn{1}{c}{Atom evidence} & 1-shot & 5-shot & 10-shot \\
\midrule
Full SADe & 26.57 & 37.04 & 38.36 \\
Target-evidence atoms removed & 23.53 & 34.85 & 36.74 \\
\bottomrule
\end{tabular}
}
\caption{Target-evidence atom removal on \cocotwenty{} box-r4. For each support example, active target-evidence atoms are ranked by a reliability-weighted target-evidence score, and weights for up to eight top-ranked atoms are zeroed; weak support, query features, router, projection rule, and prediction head remain fixed. Query mIoU (\%) is rounded to two decimals; reported drops use unrounded results.}
\label{tab:atom_removal}
\end{table}

Table~\ref{tab:ablation} compares the roles of dense evidence, SAE atom evidence, and the learned reliability estimator across controlled component combinations. Dense-only and Atom-only use $d_i$ and $a_i$; Dense+Atom applies the fixed fusion $\operatorname{clip}(d_i+1.5(a_i-d_i),0,1)$ to test whether atom evidence directly improves dense-similarity-based reliability estimates without a learned router. Dense+Router omits atom inputs, whereas full SADe uses both evidence sources. Appendix~\ref{sec:lambda-study} further analyzes the Dense+Atom fusion coefficient and shows that, on both FSS-1000 and \cocotwenty{}, this benefit persists across a broad range of coefficient values.

All variants in Table~\ref{tab:ablation} are evaluated on the same 2,400 paired box-r4 episodes, and their results show a consistent progression. Atom-only outperforms Dense-only, indicating that the sparse factor-level evidence provided by SAE atoms can support reliability estimation on its own and is more effective in this setting than dense similarity alone. In the fixed-fusion control, Dense+Atom raises mean mIoU from 31.24 to 32.55. With the learned reliability estimator, Dense+Router reaches 32.85 without atom inputs, while full SADe reaches 33.99 with both evidence sources. Together, the gains from Dense-only to Dense+Atom and from Dense+Router to full SADe show that atom evidence contributes non-redundant information about support reliability that dense similarity does not fully capture; whether incorporated through fixed fusion or the learned estimator, this information further improves downstream query segmentation.

Table~\ref{tab:atom_removal} reports a control that zeros up to eight top-scoring active target-evidence atoms per support example while holding episodes, dense evidence, and all model components fixed. Box-r4 mIoU drops by 3.04, 2.18, and 1.62 points at 1, 5, and 10 shots. These consistent drops show that the removed high-scoring target-evidence atoms make a distinct contribution to full SADe that is not recovered by dense evidence or the remaining atom inputs; this effect persists from 1 to 10 shots.

\section{Conclusion}

We introduced SADe, a predictor-agnostic support-annotation cleaner that combines dense similarity with SAE atom evidence to estimate patch reliability and produce a standard cleaned mask. Using only support-side information, SADe decouples cleaning from query inference and integrates with heterogeneous FSS predictors without altering their native query-inference paths. On \cocotwenty{}, SADe leads six of nine standalone prompt--shot settings. In paired box-family plug-in evaluation, gains span all four predictors on \cocotwenty{} and persist on \pascalfive{} without PASCAL-specific training, fitting, or checkpoint selection. Across both datasets, mean gains are larger under expanded boxes than tight boxes, consistent with suppressing the additional target-irrelevant context. The ablations further show that SAE atom evidence contributes reliability information beyond dense similarity. Overall, SADe enables support cleaning to be reused across heterogeneous FSS predictors, allowing each predictor to construct class evidence from more reliable support regions.
\bibliography{references}

\clearpage
\setcounter{secnumdepth}{2}
\appendix
\numberwithin{equation}{section}
\numberwithin{figure}{section}
\numberwithin{table}{section}
\twocolumn[
\vbox to \titlebox{%
  \vskip 0.625in minus 0.125in%
  \centering%
  {\LARGE\bfseries Technical Appendix for SADe: Sparse-Atom Support Decontamination for Few-Shot Segmentation with Weak Support Annotations\par}%
  \vfill%
}%
]
\section{Appendix Purpose and Scope}
\label{sec:supp-scope}

This appendix extends the main paper in three ways. First, it provides the implementation details needed to reproduce SADe, including robust score standardization, router architecture and initialization, training, and model selection, together with support-side latency and CUDA memory measurements. Second, it characterizes the sensitivity of the fixed Dense+Atom control to the residual coefficient $\lambda_{\mathrm{res}}$, testing whether atom evidence continues to improve dense-similarity-based reliability estimates across coefficient values. Third, it adds a standalone evaluation on \pascalfive{}, extends the standalone and paired plug-in evaluations to \lvisninetytwo{} and to remote-sensing and medical-imaging domains, and compares generic and domain-related unlabeled image sources.

The sensitivity and generalization analyses address three questions. First, they examine whether atom evidence improves dense-similarity-based reliability estimates across a broad range of residual weights in the fixed Dense+Atom control. Second, they test whether SADe's support-cleaning gains and predictor-agnostic reuse across heterogeneous downstream models extend to broader datasets and visual domains. Third, they examine whether the same frozen router remains effective when the SAE dictionary, dense-branch PCA mean and principal-component basis, and atom activation-frequency estimates are obtained from different unlabeled image sources. To examine the third question, we first apply the components obtained from generic COCO imagery directly to aerial remote sensing and medical endoscopy, and then evaluate the corresponding components independently obtained from domain-related imagery with the same frozen router.

\section{Numerical Details and Router Training}
\label{sec:implementation-details}

The reliability router performs patch-level reliability estimation and episode-level mixing. For support patch $i$, the local feature $e_i$ summarizes calibrated dense and atom evidence, while the episode summary $E$ characterizes the evidence distribution of the current support episode. The router predicts patch reliability $R_\theta(e_i,E)$ from $(e_i,E)$ and an episode-level mixing weight $\alpha(E)$ from $E$, combining the learned reliability with dense confidence $d_i$. The remainder of this section specifies robust score standardization, router architecture and initialization, and the training and model-selection procedure.

\noindent\textbf{Robust score standardization.}
The scales of dense and atom scores can vary across episodes and can also differ between the two evidence branches. They are therefore standardized independently over valid support patches before constructing $e_i$ and $E$. Following the patch-validity rule in the main paper, let $s_V$ contain the scores on valid support patches for either raw score vector $s\in\{s^d,s^a\}$, and let $Q_p(s_V)$ denote its empirical $p$-quantile. Both the center and scale are estimated from $s_V$. The center is its median:
\begin{equation}
m_s=Q_{0.5}(s_V).
\label{eq:robust-center}
\end{equation}
The IQR-based scale is
\begin{equation}
\widetilde{\sigma}_s
=
\frac{Q_{0.75}(s_V)-Q_{0.25}(s_V)}{1.349}.
\label{eq:robust-scale}
\end{equation}
Because $1.349$ is the interquartile range of a standard normal variable, the divisor maps the IQR estimate to the standard-deviation scale under normality. We set $\sigma_s=\widetilde{\sigma}_s$ when this estimate is finite and at least $10^{-6}$. Otherwise, $\sigma_s$ is the population standard deviation of $s_V$, with a final fallback of $\sigma_s=1$ if that estimate is also non-finite or below the threshold. The standardized score is
\begin{equation}
\zeta(s_i)=\frac{s_i-m_s}{\sigma_s}.
\label{eq:robust-transform}
\end{equation}

\noindent\textbf{Router architecture and initialization.}
The standardized scores contribute to the seven-dimensional patch feature $e_i$ and the eight-dimensional episode summary $E$. The episode MLP maps $E$ to a 96-dimensional context representation. The patch MLP concatenates this representation with $e_i$ and produces the reliability logit $\ell_i$, whereas the scalar mixing branch combines the same representation with $E$ to produce the mixing logit $z_\alpha$. With $\sigma$ denoting the sigmoid, the two branch outputs are $R_\theta(e_i,E)=\sigma(\ell_i)$ and $\alpha(E)=\sigma(z_\alpha)$. Table~\ref{tab:router-architecture} gives the exact dimensions and fixed constants.

\begin{table*}[!t]
\centering
{\small
\setlength{\tabcolsep}{5.0pt}
\begin{tabular}{@{}V{0.35\textwidth}V{0.56\textwidth}@{}}
\toprule
\multicolumn{1}{c}{Component} & \multicolumn{1}{c}{Setting} \\
\midrule
Episode MLP & LayerNorm(8), $8\!\rightarrow\!96\!\rightarrow\!96$ MLP \\
Patch MLP & LayerNorm(103), $103\!\rightarrow\!96\!\rightarrow\!96\!\rightarrow\!1$ MLP \\
Scalar mixing branch & LayerNorm(104), $104\!\rightarrow\!96\!\rightarrow\!48\!\rightarrow\!1$ MLP \\
Activation and regularization & GELU; dropout $0.05$; gradient clipping at $1.0$ \\
Episode-salient atoms & $K_a=128$ \\
Loss, mixing-target, and initialization constants & $\lambda_s=1.0$; $\lambda_\alpha=0.25$; $c_\alpha=0.52$; $\beta_\alpha=8$; $\alpha^\star\in[0.05,0.95]$; $\alpha_0=0.35$ \\
\bottomrule
\end{tabular}
}
\caption{Architecture and fixed constants used by the reliability router described in the main paper.}
\label{tab:router-architecture}
\end{table*}

To make the initial predictions independent of as-yet untrained patch and episode features, the output-layer weights of both the patch MLP and scalar mixing branch are initialized to zero. Setting the patch-MLP output bias to zero gives $\ell_i=0$ and therefore $R_\theta=\sigma(0)=0.5$ for every patch. For the scalar branch, setting the output bias to $\operatorname{logit}(0.35)$, where $\operatorname{logit}(p)=\ln[p/(1-p)]$, gives $z_\alpha=\operatorname{logit}(0.35)$ and therefore $\alpha(E)=\sigma(\operatorname{logit}(0.35))=0.35$ for every episode. Training then learns patch-dependent reliability and episode-dependent mixing from this common starting point.

\noindent\textbf{Training and model selection.}
From the FSS-1000 class pool, we use 700 classes for router optimization and two additional, mutually disjoint sets of 120 classes for validation and held-out analysis. The validation set is used for checkpoint selection and the validation-side sensitivity curve, whereas the held-out set is used only for the test-side sensitivity curve. Both sets are sampled once with fixed seeds and remain unchanged throughout the experiments.

The router is trained on weak-support episodes synthesized from clean FSS-1000 support masks, exposing it to different patterns of target-irrelevant context and coarse boundaries. Within each synthetic weak-support region, the clean mask supplies the patch-reliability labels. Their positive fraction defines weak-support purity $\rho$, from which the episode-level mixing target $\alpha^\star(\rho)$ is computed using the mapping defined in the main paper. At each optimization step, one of seven weak-support forms is sampled uniformly: three box masks (tight, box-r2, and box-r4), two coarse masks, and two dilation-based training augmentations. For the coarse masks, the support-token grid is partitioned into non-overlapping $4\times4$ or $8\times8$ patch blocks. A block is marked as foreground if it contains at least one foreground patch from the clean support mask, with boundary blocks restricted to the valid grid. Both dilation-based forms start from the patch-level foreground mask derived from the clean support mask. The first expands each foreground patch to a square neighborhood with a radius of two grid cells. The second applies the same dilation, samples without replacement from patches labeled as background by the clean support annotation, and takes the union of the sampled patches and the dilated mask. The sample count is obtained by rounding 50\% of the dilated-mask size and is capped by the number of available background patches. The router is optimized for 800 steps with AdamW using a learning rate of $3\times10^{-4}$ and weight decay of $10^{-4}$.

Every 80 steps, the current router is evaluated on one fixed support-query episode from each validation class~\citep{li2020fss1000} under the three box and two coarse-mask forms. Checkpoint selection uses only these five validation forms; the two dilation-based forms serve only as training augmentations. The selection criterion balances average improvement against regressions on individual validation forms. Let $\Delta_g$ denote the mean paired query-Dice gain over Dense-only for form $g$, and let $\mathcal{G}$ contain these five validation forms. The selection score is
\begin{equation}
\frac{1}{|\mathcal{G}|}\sum_{g\in\mathcal{G}}\Delta_g
-0.5\frac{1}{|\mathcal{G}|}\sum_{g\in\mathcal{G}}\max(-\Delta_g,0).
\label{eq:checkpoint-selection-score}
\end{equation}
The first term rewards mean improvement across the validation forms, while the second penalizes forms with negative mean gains. This criterion is used to select the final reliability router, whose parameters are then frozen for all target evaluations.

\section{Support-Side Runtime and Memory}
\label{sec:efficiency}

Because SADe cleans each support annotation once and reuses the resulting mask across all queries that share the support set, the benchmark measures support-side preprocessing alone and excludes downstream query prediction.

\noindent\textbf{Protocol.}
The benchmark comprises 200 \cocotwenty{} tight-box support cases sampled round-robin from the fixed episode manifest to cover all 80 classes. All measurements use batch size one, FP32, and direct resizing to $512\times512$ without padding. After loading the resident models required by each execution regime, we perform 600 warm-up calls and three timed passes over the complete case set, yielding 600 CUDA-synchronized observations. Inputs are preloaded; model loading, disk I/O, image decoding, and preprocessing are excluded from the timing. Memory is reported as both total peak CUDA allocation and the increment above the resident state. Measurements use an NVIDIA GeForce RTX 5090, an Intel Core i9-14900KF, four CPU threads, PyTorch 2.11.0, and CUDA 13.0.

\begin{table*}[!t]
\centering
{\small
\setlength{\tabcolsep}{3.0pt}
\begin{tabular}{@{}V{0.29\textwidth}M{0.14\textwidth}M{0.09\textwidth}M{0.09\textwidth}M{0.13\textwidth}M{0.13\textwidth}@{}}
\toprule
\multicolumn{1}{c}{Available input} & Mean $\pm$ SD (ms) & Median (ms) & P95 (ms) & Peak allocated (MiB) & Incremental peak (MiB) \\
\midrule
DINOv3 features + atom activations & $16.07 \pm 1.57$ & 16.07 & 18.68 & 10.51 & 1.21 \\
DINOv3 features & $17.94 \pm 1.69$ & 18.01 & 20.54 & 126.27 & 68.90 \\
Preprocessed support image & $43.55 \pm 6.50$ & 42.77 & 56.12 & 453.18 & 68.90 \\
\bottomrule
\end{tabular}
}
\caption{Support-side latency and CUDA memory under three input regimes. SADe cleaning comprises evidence construction, router inference, and mask projection. P95 denotes the 95th-percentile latency. Total peak includes the resident models required by each regime; incremental peak measures the additional allocation during execution. Each regime is measured in an independent process using 600 CUDA-synchronized observations.}
\label{tab:efficiency}
\end{table*}

Table~\ref{tab:efficiency} separates the three execution regimes by the inputs available at the start of cleaning. With DINOv3~\citep{simeoni2026dinov3} features and SAE atom activations cached, evidence construction, router inference, and mask projection have a median latency of 16.07\,ms, a P95 latency of 18.68\,ms, and an incremental peak allocation of 1.21\,MiB. Starting from cached DINOv3 features raises the median to 18.01\,ms, indicating that online SAE encoding adds approximately 1.94\,ms. Starting from a preprocessed support image raises the median to 42.77\,ms and the P95 latency to 56.12\,ms, showing that most of the remaining latency comes from DINOv3 feature extraction.

The latter two regimes have the same incremental peak of 68.90\,MiB. The 453.18\,MiB total peak of the complete path therefore primarily reflects the resident DINOv3 and SAE models rather than temporary allocation added by each cleaning call. Comparing the three regimes isolates the costs of feature extraction, SAE encoding, and SADe cleaning. When support features and atom activations are cached, the median cleaning latency and incremental peak allocation decrease to 16.07\,ms and 1.21\,MiB, respectively, showing that cached support representations substantially reduce the support-side latency and incremental memory required to produce a cleaned mask.

\section{Dense+Atom Coefficient Sensitivity}
\label{sec:lambda-study}

The Dense+Atom control in the main paper uses a fixed residual combination to test whether atom evidence directly improves dense-similarity-based reliability estimates without a learned router. We systematically sweep $\lambda_{\mathrm{res}}$ on FSS-1000 and \cocotwenty{} to characterize how the complementary contribution of atom evidence to these reliability estimates varies with the residual-fusion weight and to assess its stability over a broad coefficient range. The fixed-fusion score is
\begin{equation}
q_i(\lambda_{\mathrm{res}})=
\operatorname{clip}\!\left(d_i+\lambda_{\mathrm{res}}(a_i-d_i),0,1\right).
\label{eq:dense-atom-residual}
\end{equation}
Here, $\lambda_{\mathrm{res}}=0$ recovers Dense-only and $\lambda_{\mathrm{res}}=1$ recovers Atom-only; values above one further amplify the residual $a_i-d_i$ of the atom score relative to the dense score.

\subsection{Sensitivity Protocol}

We sweep $\lambda_{\mathrm{res}}\in[0,4]$ in increments of 0.25. To test whether the complementary contribution of atom evidence generalizes across classes, we estimate separate sensitivity curves on the fixed, class-disjoint validation and test partitions introduced above. The same coefficient grid is used for both partitions, and all coefficients within a partition share identical retained episodes. Performance is reported as episode-mean query Dice. For a predicted query mask $P$ and ground-truth mask $Y$, Dice is defined as
\begin{equation}
\operatorname{Dice}(P,Y)=\frac{2|P\cap Y|}{|P|+|Y|}.
\label{eq:query-dice}
\end{equation}

\begin{figure*}[!t]
\centering
\includegraphics[width=\textwidth]{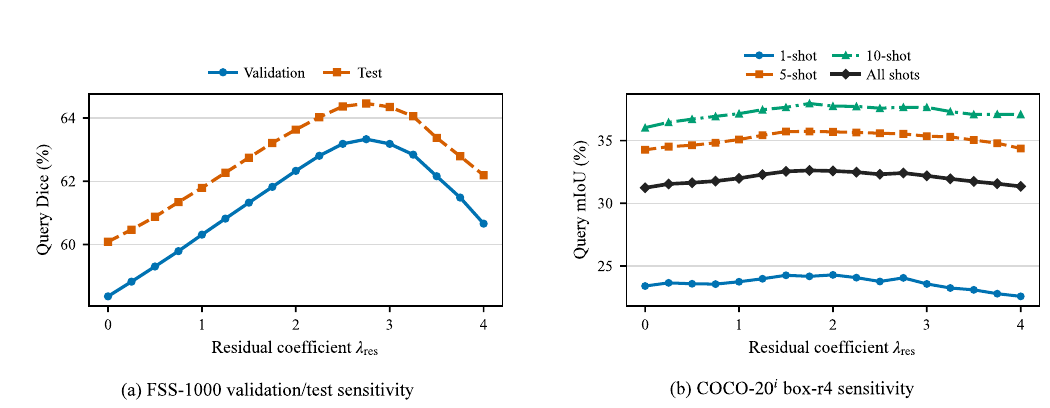}
\caption{Residual-coefficient sensitivity of the fixed Dense+Atom control. (a) Episode-mean query Dice on the fixed, class-disjoint FSS-1000 validation and test partitions. (b) Query mIoU on \cocotwenty{} box-r4 for 1/5/10-shot settings and their aggregate. Within each setting, all coefficients use identical episodes with cached scores, projection, and query prediction held fixed.}
\label{fig:lambda-sensitivity}
\end{figure*}

On \cocotwenty{} box-r4, we evaluate the same grid on all 2,400 episodes while holding the episodes, cached dense and atom scores, projection rule, and query-prediction procedure fixed. We report query mIoU for each shot and across shots to evaluate coefficient sensitivity under the main-paper protocol. Figure~\ref{fig:lambda-sensitivity} summarizes both sweeps, and Table~\ref{tab:lambda-full-sweep} reports the complete \cocotwenty{} results.

\begin{table}[!t]
\centering
\small
\setlength{\tabcolsep}{2.6pt}
\begin{tabular*}{0.86\columnwidth}{@{\extracolsep{\fill}}crrrr@{}}
\toprule
$\lambda_{\mathrm{res}}$ & \multicolumn{1}{c}{1-shot} & \multicolumn{1}{c}{5-shot} & \multicolumn{1}{c}{10-shot} & \multicolumn{1}{c}{Mean} \\
\midrule
0 & 23.41 & 34.27 & 36.03 & 31.24 \\
0.25 & 23.66 & 34.52 & 36.46 & 31.55 \\
0.5 & 23.59 & 34.63 & 36.71 & 31.64 \\
0.75 & 23.56 & 34.83 & 36.93 & 31.77 \\
1 & 23.75 & 35.09 & 37.15 & 32.00 \\
1.25 & 23.99 & 35.42 & 37.47 & 32.29 \\
1.5 & 24.26 & 35.72 & 37.65 & 32.55 \\
1.75 & 24.18 & \textbf{35.73} & \textbf{37.96} & \textbf{32.62} \\
2 & \textbf{24.30} & 35.70 & 37.75 & 32.58 \\
2.25 & 24.07 & 35.65 & 37.73 & 32.49 \\
2.5 & 23.77 & 35.59 & 37.58 & 32.31 \\
2.75 & 24.06 & 35.54 & 37.66 & 32.42 \\
3 & 23.57 & 35.35 & 37.65 & 32.19 \\
3.25 & 23.25 & 35.30 & 37.32 & 31.95 \\
3.5 & 23.10 & 35.04 & 37.08 & 31.74 \\
3.75 & 22.80 & 34.79 & 37.10 & 31.56 \\
4 & 22.58 & 34.37 & 37.09 & 31.35 \\
\bottomrule
\end{tabular*}
\caption{Complete $\lambda_{\mathrm{res}}$ sensitivity sweep on \cocotwenty{} box-r4 (query mIoU, \%). Each shot contains 800 episodes; the Mean column aggregates all 2,400 episodes across the three shots. The best value in each column is bolded.}
\label{tab:lambda-full-sweep}
\end{table}

\subsection{Sensitivity Results}

On FSS-1000, validation and test query Dice exhibit the same coefficient-dependent trend: both rise from $\lambda_{\mathrm{res}}=0$, remain close to their maxima over $[2.5,3.0]$, and decline at larger coefficients. Within this interval, validation query Dice ranges from $63.18$ to $63.33$, while test query Dice ranges from $64.35$ to $64.46$. The agreement between the class-disjoint splits shows that the complementary contribution of atom evidence is stable across classes.

Across the 2,400 \cocotwenty{} box-r4 episodes, aggregate query mIoU rises above the Dense-only value of $31.24$ and remains between $32.29$ and $32.62$ throughout $\lambda_{\mathrm{res}}\in[1.25,2.75]$. The main-paper setting $\lambda_{\mathrm{res}}=\PaperResidualLambda$ reaches $32.55$ mIoU and lies within this high-performing interval; moreover, every nonzero coefficient outperforms Dense-only. Together, the two analyses show that atom evidence contributes support-reliability information not fully captured by dense similarity and that this contribution persists across residual coefficients rather than depending on a narrowly tuned value.

\section{Extended Evaluation Protocol}
\label{sec:additional-settings}

The extended evaluations preserve the matched ProMi~\citep{chiaroni2025promi} standalone comparison and paired plug-in protocol of the main paper. Building on this protocol, we add a \pascalfive{} standalone comparison under the shared ProMi query head and evaluate broader natural-image categories on \lvisninetytwo{}, aerial remote-sensing imagery on \isaidfive{}, and cross-dataset medical endoscopy with Kvasir-SEG support and CVC-ClinicDB queries.

\cocotwenty{}~\citep{lin2014coco,nguyen2019feature} and \pascalfive{}~\citep{shaban2017oneshot} use their standard class splits and a fixed, class-balanced episode manifest for each prompt--shot setting.

\noindent\textbf{Evaluation Metric and Aggregation.}
For each episode included in the evaluation, let $P$ and $Y$ denote the binary predicted and ground-truth query masks, respectively, and define
\begin{equation}
\operatorname{IoU}(P,Y)=\frac{|P\cap Y|}{|P\cup Y|}.
\label{eq:query-iou}
\end{equation}
For benchmarks organized into folds, we first compute the mean episode-level IoU within each fold and then take an equally weighted mean across folds. For settings without folds, episode-level IoUs are averaged directly over the fixed manifest. Tables report query mIoU in percent; each cross-shot mean is the arithmetic mean of the corresponding unrounded 1-, 5-, and 10-shot mIoU values. All differences reported below are computed from the corresponding unrounded results.

\subsection{Predictor-Specific Preprocessing}

All four plug-in predictors---FSSDINO~\citep{zakir2026fssdino}, INSID3~\citep{cuttano2026insid3}, SANSA~\citep{cuttano2025sansa}, and FSS-SAM3~\citep{tsai2026fsssam3}---retain the model parameters and native support interfaces used in the main paper. Table~\ref{tab:native-interfaces} summarizes the predictor-specific preprocessing used across the extended evaluation grid.

\begin{table*}[!t]
\centering
{\small
\setlength{\tabcolsep}{4.0pt}
\begin{tabular}{@{}V{0.22\textwidth}V{0.68\textwidth}@{}}
\toprule
\multicolumn{1}{c}{Downstream predictor} & \multicolumn{1}{c}{Preprocessing} \\
\midrule
FSSDINO and SANSA & Images and masks are resized directly to $512\times512$ without padding \\
INSID3 & Support masks are mapped to original-image coordinates; inference uses the native 1024-pixel resolution \\
FSS-SAM3 & Original images are arranged on a $1008\times1008$ collage canvas \\
\bottomrule
\end{tabular}
}
\caption{Predictor-specific preprocessing for the extended plug-in evaluations.}
\label{tab:native-interfaces}
\end{table*}

\subsection{Episode Construction and Unlabeled Sources}
\label{sec:datasets-sources}

\noindent\textbf{Episode construction.}
\emph{\lvisninetytwo{}}~\citep{gupta2019lvis}. For each $s\in\{1,5,10\}$, a candidate class must contain at least $s+1$ annotated images, allowing $s$ distinct support images and a query image not used as support to be selected. Preserving the official class order, we retain the longest prefix of the candidate sequence whose size is divisible by ten and assign its classes round-robin to ten equally sized folds. Images are then filtered according to the validity of their target-class semantic masks, and fixed manifests are constructed from the class--shot pairs that satisfy the sampling requirement.

\noindent\emph{\isaidfive{}.} Episodes follow the official split lists introduced with the benchmark~\citep{yao2022scaleaware}, and the manifests are fixed before evaluation.

\noindent\emph{Medical endoscopy.} We use fixed cross-dataset manifests with support images from Kvasir-SEG~\citep{jha2020kvasir} and query images from CVC-ClinicDB~\citep{bernal2015wmdova}.

\noindent\textbf{Unlabeled image sources.}
The source labels in the result tables refer to COCO train2014, FSS-1000 images, the iSAID~\citep{zamir2019isaid} training set, or the Kvasir-SEG training set. For each source, the same unlabeled image pool is used to train the SAE, fit the dense-branch PCA mean and principal-component basis, and estimate the activation frequencies used for high-frequency atom exclusion. All SAEs are trained independently on their respective unlabeled features and remain frozen during subsequent use. The COCO, iSAID, and Kvasir-SEG sources use 16k-atom SAEs with a BatchTopK~\citep{bussmann2024batchtopk} budget of 32; the FSS-1000 source uses the 8k-atom, budget-24 SAE that supplies atom evidence during router training.

\section{Generalization Across Datasets and Visual Domains}
\label{sec:transfer-results}

The standalone results compare support-mask construction under a shared ProMi query head, while the plug-in results test whether SADe remains reusable across the four downstream predictors on the added datasets and visual domains. The unlabeled-source comparisons then evaluate the same frozen router with SAE dictionaries, dense-branch PCA means and principal-component bases, and atom activation-frequency estimates obtained from either generic or domain-related unlabeled imagery.

\subsection{Standalone Generalization}

The standalone results cover six dataset--source configurations across natural imagery (Tables~\ref{tab:appendix-standalone-pascal_coco_sae}--\ref{tab:appendix-standalone-lvis_fss1000_sae}), aerial remote sensing (Tables~\ref{tab:appendix-standalone-isaid_coco_sae}--\ref{tab:appendix-standalone-isaid_domain_sae}), and medical endoscopy (Tables~\ref{tab:appendix-standalone-polyp_coco_sae}--\ref{tab:appendix-standalone-polyp_domain_sae}). After averaging over the nine prompt--shot settings in each dataset--source configuration, SADe achieves a higher mean query mIoU than raw-mask ProMi and SAM3-guided ProMi~\citep{carion2025sam3segmentconcepts} in all six configurations. Relative to raw-mask ProMi, the gains are $\StandalonePascalMeanDelta{}$ and $\StandaloneLvisMeanDelta{}$ mIoU points on \pascalfive{} and \lvisninetytwo{}, $\StandaloneISaidCocoMeanDelta{}$ and $\StandaloneISaidDomainMeanDelta{}$ points on \isaidfive{} with COCO and iSAID sources, and $\StandalonePolypCocoMeanDelta{}$ and $\StandalonePolypDomainMeanDelta{}$ points in medical endoscopy with COCO and Kvasir-SEG sources. Under the shared ProMi query-prediction path, SADe outperforms raw-mask ProMi in all six natural-image, aerial remote-sensing, and medical-endoscopy configurations.

\begin{table}[!tb]
\centering
\small
\setlength{\tabcolsep}{1mm}
\begin{tabular}{@{}clrrrr@{}}
\toprule
Method & Prompt & 1-shot & 5-shot & 10-shot & Avg. \\
\midrule
\multirow{3}{*}{ProMi} & box & \textbf{55.45} & \textbf{67.44} & \textbf{69.31} & \textbf{64.07} \\
 & box-r2 & 47.00 & 61.79 & \textbf{66.01} & 58.27 \\
 & box-r4 & 34.57 & 50.25 & 56.20 & 47.01 \\
\midrule
\multirow{3}{*}{\shortstack{SAM3-guided\\ProMi}} & box & 51.90 & 58.57 & 60.29 & 56.92 \\
 & box-r2 & 48.62 & 55.01 & 55.06 & 52.90 \\
 & box-r4 & 37.68 & 40.65 & 42.27 & 40.20 \\
\midrule
\multirow{3}{*}{SADe} & box & 45.97 & 59.14 & 60.92 & 55.35 \\
 & box-r2 & \textbf{53.00} & \textbf{63.75} & 65.42 & \textbf{60.72} \\
 & box-r4 & \textbf{50.28} & \textbf{62.82} & \textbf{65.35} & \textbf{59.49} \\
\bottomrule
\end{tabular}
\caption{Standalone query mIoU (\%) on \pascalfive{} (SADe unlabeled source: COCO). All three methods use the same ProMi query-prediction path and differ only in support-mask construction. Bold indicates the best method in each prompt--shot cell and prompt-wise cross-shot mean.}
\label{tab:appendix-standalone-pascal_coco_sae}
\end{table}

\begin{table}[!tb]
\centering
\small
\setlength{\tabcolsep}{1mm}
\begin{tabular}{@{}clrrrr@{}}
\toprule
Method & Prompt & 1-shot & 5-shot & 10-shot & Avg. \\
\midrule
\multirow{3}{*}{ProMi} & box & 23.28 & 28.78 & 31.39 & 27.81 \\
 & box-r2 & 15.93 & 22.97 & 26.28 & 21.73 \\
 & box-r4 & 11.59 & 17.82 & 21.66 & 17.02 \\
\midrule
\multirow{3}{*}{\shortstack{SAM3-guided\\ProMi}} & box & \textbf{24.85} & \textbf{30.45} & \textbf{33.09} & \textbf{29.47} \\
 & box-r2 & 13.88 & 18.36 & 20.37 & 17.54 \\
 & box-r4 & 8.40 & 11.08 & 12.62 & 10.70 \\
\midrule
\multirow{3}{*}{SADe} & box & 22.24 & 29.14 & 31.13 & 27.50 \\
 & box-r2 & \textbf{18.97} & \textbf{26.64} & \textbf{30.16} & \textbf{25.26} \\
 & box-r4 & \textbf{15.00} & \textbf{23.05} & \textbf{27.66} & \textbf{21.90} \\
\bottomrule
\end{tabular}
\caption{Standalone query mIoU (\%) on \lvisninetytwo{} (SADe unlabeled source: FSS-1000). All three methods use the same ProMi query-prediction path and differ only in support-mask construction. Bold indicates the best method in each prompt--shot cell and prompt-wise cross-shot mean.}
\label{tab:appendix-standalone-lvis_fss1000_sae}
\end{table}

\begin{table}[!tb]
\centering
\small
\setlength{\tabcolsep}{1mm}
\begin{tabular}{@{}clrrrr@{}}
\toprule
Method & Prompt & 1-shot & 5-shot & 10-shot & Avg. \\
\midrule
\multirow{3}{*}{ProMi} & box & \textbf{30.89} & \textbf{38.82} & \textbf{40.29} & \textbf{36.67} \\
 & box-r2 & 28.56 & 35.87 & 38.65 & 34.36 \\
 & box-r4 & 25.68 & 32.11 & 36.96 & 31.59 \\
\midrule
\multirow{3}{*}{\shortstack{SAM3-guided\\ProMi}} & box & 27.84 & 33.91 & 30.53 & 30.76 \\
 & box-r2 & 25.59 & 28.99 & 30.15 & 28.24 \\
 & box-r4 & 21.99 & 22.50 & 25.48 & 23.33 \\
\midrule
\multirow{3}{*}{SADe} & box & 26.87 & 38.81 & 39.50 & 35.06 \\
 & box-r2 & \textbf{28.96} & \textbf{39.95} & \textbf{40.45} & \textbf{36.45} \\
 & box-r4 & \textbf{28.50} & \textbf{37.79} & \textbf{40.58} & \textbf{35.62} \\
\bottomrule
\end{tabular}
\caption{Standalone query mIoU (\%) on \isaidfive{} (SADe unlabeled source: COCO). All three methods use the same ProMi query-prediction path and differ only in support-mask construction. Bold indicates the best method in each prompt--shot cell and prompt-wise cross-shot mean.}
\label{tab:appendix-standalone-isaid_coco_sae}
\end{table}

\begin{table}[!tb]
\centering
\small
\setlength{\tabcolsep}{1mm}
\begin{tabular}{@{}clrrrr@{}}
\toprule
Method & Prompt & 1-shot & 5-shot & 10-shot & Avg. \\
\midrule
\multirow{3}{*}{ProMi} & box & \textbf{30.89} & 38.82 & \textbf{40.29} & \textbf{36.67} \\
 & box-r2 & 28.56 & 35.87 & 38.65 & 34.36 \\
 & box-r4 & 25.68 & 32.11 & 36.96 & 31.59 \\
\midrule
\multirow{3}{*}{\shortstack{SAM3-guided\\ProMi}} & box & 27.84 & 33.91 & 30.53 & 30.76 \\
 & box-r2 & 25.59 & 28.99 & 30.15 & 28.24 \\
 & box-r4 & 21.99 & 22.50 & 25.48 & 23.33 \\
\midrule
\multirow{3}{*}{SADe} & box & 26.93 & \textbf{39.37} & 39.67 & 35.32 \\
 & box-r2 & \textbf{29.45} & \textbf{40.18} & \textbf{40.72} & \textbf{36.78} \\
 & box-r4 & \textbf{29.33} & \textbf{38.28} & \textbf{40.68} & \textbf{36.09} \\
\bottomrule
\end{tabular}
\caption{Standalone query mIoU (\%) on \isaidfive{} (SADe unlabeled source: iSAID). All three methods use the same ProMi query-prediction path and differ only in support-mask construction. Bold indicates the best method in each prompt--shot cell and prompt-wise cross-shot mean.}
\label{tab:appendix-standalone-isaid_domain_sae}
\end{table}

Prompt-wise averages show a consistent SADe advantage under box expansion. The leading method under tight boxes varies across configurations; in contrast, SADe attains the highest cross-shot mean query mIoU for both box-r2 and box-r4 in every configuration, ranking first in all $6\times2=12$ expanded-box comparisons. Moreover, its gain over raw-mask ProMi increases from box-r2 to box-r4 in every configuration. Thus, the increase in SADe's gain from box-r2 to box-r4 observed in the main paper also appears in the broader natural-image, aerial remote-sensing, and medical-endoscopy settings.

The matched SAM3-guided ProMi experiment compares two support-mask construction strategies under expanded boxes: SADe's reliability-guided cleaning and generic box-prompt segmentation. The two methods share the ProMi query-prediction path but construct their support masks differently; SADe achieves a higher cross-shot mean query mIoU in all 12 comparisons, showing that its cleaned masks provide more effective support inputs than masks obtained from SAM3 box prompting. Together with the main-paper ablations, these cross-domain results indicate that combining dense similarity with foreground/complement atom evidence enables the reliability router to identify target-relevant reliable patches and suppress context introduced by expanded boxes.

\begin{table}[!tb]
\centering
\small
\setlength{\tabcolsep}{1mm}
\begin{tabular}{@{}clrrrr@{}}
\toprule
Method & Prompt & 1-shot & 5-shot & 10-shot & Avg. \\
\midrule
\multirow{3}{*}{ProMi} & box & \textbf{41.80} & \textbf{46.56} & \textbf{50.34} & \textbf{46.23} \\
 & box-r2 & 39.06 & 43.44 & \textbf{52.72} & 45.07 \\
 & box-r4 & 28.64 & 38.76 & 46.65 & 38.02 \\
\midrule
\multirow{3}{*}{\shortstack{SAM3-guided\\ProMi}} & box & 36.01 & 45.78 & 47.29 & 43.03 \\
 & box-r2 & 38.21 & 44.31 & 50.74 & 44.42 \\
 & box-r4 & 32.01 & 36.86 & 47.23 & 38.70 \\
\midrule
\multirow{3}{*}{SADe} & box & 27.23 & 45.27 & 44.84 & 39.11 \\
 & box-r2 & \textbf{42.37} & \textbf{49.39} & 49.79 & \textbf{47.18} \\
 & box-r4 & \textbf{36.32} & \textbf{47.03} & \textbf{50.23} & \textbf{44.53} \\
\bottomrule
\end{tabular}
\caption{Standalone segmentation results with Kvasir-SEG support and CVC-ClinicDB queries (query mIoU, \%). SADe uses COCO as its unlabeled source. Methods share the ProMi query-prediction path and differ only in support-mask construction. Bold marks the best value in each prompt--shot cell and cross-shot mean.}
\label{tab:appendix-standalone-polyp_coco_sae}
\end{table}

\begin{table}[!tb]
\centering
\small
\setlength{\tabcolsep}{1mm}
\begin{tabular}{@{}clrrrr@{}}
\toprule
Method & Prompt & 1-shot & 5-shot & 10-shot & Avg. \\
\midrule
\multirow{3}{*}{ProMi} & box & \textbf{41.80} & \textbf{46.56} & \textbf{50.34} & \textbf{46.23} \\
 & box-r2 & 39.06 & 43.44 & \textbf{52.72} & 45.07 \\
 & box-r4 & 28.64 & 38.76 & 46.65 & 38.02 \\
\midrule
\multirow{3}{*}{\shortstack{SAM3-guided\\ProMi}} & box & 36.01 & 45.78 & 47.29 & 43.03 \\
 & box-r2 & 38.21 & 44.31 & 50.74 & 44.42 \\
 & box-r4 & 32.01 & 36.86 & 47.23 & 38.70 \\
\midrule
\multirow{3}{*}{SADe} & box & 26.81 & 44.52 & 44.53 & 38.62 \\
 & box-r2 & \textbf{43.49} & \textbf{48.24} & 49.00 & \textbf{46.91} \\
 & box-r4 & \textbf{39.35} & \textbf{49.77} & \textbf{51.39} & \textbf{46.84} \\
\bottomrule
\end{tabular}
\caption{Standalone segmentation results with Kvasir-SEG support and CVC-ClinicDB queries (query mIoU, \%). SADe uses Kvasir-SEG as its unlabeled source. Methods share the ProMi query-prediction path and differ only in support-mask construction. Bold marks the best value in each prompt--shot cell and cross-shot mean.}
\label{tab:appendix-standalone-polyp_domain_sae}
\end{table}

\subsection{Plug-In Generalization}

Tables~\ref{tab:appendix-full-lvis_fss1000_sae}, \ref{tab:appendix-full-isaid_coco_sae}--\ref{tab:appendix-full-isaid_domain_sae}, and~\ref{tab:appendix-full-polyp_coco_sae}--\ref{tab:appendix-full-polyp_domain_sae} report five added dataset--source configurations: \lvisninetytwo{}, two \isaidfive{} settings, and two medical-endoscopy settings. Each configuration covers four predictors, three box-family prompts, and three shot settings, yielding 36 paired Raw/SADe cells. SADe improves the mean query mIoU in all five configurations and outperforms Raw in at least \PluginSettingMinWins{}/36 cells in each; all \PluginLVISWins{} cells improve on \lvisninetytwo{}. To summarize transfer beyond the \cocotwenty{} target benchmark, we combine these five configurations with the \pascalfive{} setting from the main paper. Across the resulting six dataset--source configurations, SADe improves \AppendixPluginWins{} of \AppendixPluginCells{} paired cells, with an overall mean gain of \AppendixPluginMeanDelta{} mIoU. Thus, the plug-in gains observed on \cocotwenty{} also hold on \pascalfive{}, \lvisninetytwo{}, \isaidfive{}, and the Kvasir-SEG$\rightarrow$CVC-ClinicDB endoscopy task.

\begin{table}[!tb]
\centering
\small
\setlength{\tabcolsep}{1mm}
\begin{tabular}{@{}cc|rr|rr|rr@{}}
\toprule
\multirow[c]{2}{*}[-0.45ex]{Model} & \multirow[c]{2}{*}[-0.45ex]{Prompt} & \multicolumn{2}{c|}{1-shot} & \multicolumn{2}{c|}{5-shot} & \multicolumn{2}{c}{10-shot} \\
\cmidrule(lr){3-4}\cmidrule(lr){5-6}\cmidrule(l){7-8}
 & & \multicolumn{1}{c}{Raw} & \multicolumn{1}{c|}{SADe} & \multicolumn{1}{c}{Raw} & \multicolumn{1}{c|}{SADe} & \multicolumn{1}{c}{Raw} & \multicolumn{1}{c}{SADe} \\
\midrule
\multirow{3}{*}{FSSDINO} & box & 21.62 & \textbf{25.51} & 26.34 & \textbf{31.17} & 27.26 & \textbf{32.15} \\
 & box-r2 & 15.13 & \textbf{19.82} & 18.34 & \textbf{26.54} & 19.53 & \textbf{28.56} \\
 & box-r4 & 12.27 & \textbf{16.63} & 13.94 & \textbf{23.06} & 14.42 & \textbf{24.79} \\
\midrule
\multirow{3}{*}{INSID3} & box & 24.42 & \textbf{26.30} & 29.56 & \textbf{31.73} & 30.80 & \textbf{34.40} \\
 & box-r2 & 17.71 & \textbf{21.96} & 20.91 & \textbf{27.89} & 21.52 & \textbf{30.72} \\
 & box-r4 & 14.19 & \textbf{18.98} & 15.43 & \textbf{24.53} & 15.34 & \textbf{27.15} \\
\midrule
\multirow{3}{*}{SANSA} & box & 25.36 & \textbf{27.61} & 28.40 & \textbf{33.58} & 29.78 & \textbf{35.77} \\
 & box-r2 & 16.42 & \textbf{20.87} & 17.27 & \textbf{24.61} & 18.98 & \textbf{25.92} \\
 & box-r4 & 12.79 & \textbf{17.34} & 14.46 & \textbf{21.72} & 14.47 & \textbf{21.45} \\
\midrule
\multirow{3}{*}{FSS-SAM3} & box & 43.90 & \textbf{44.54} & 52.66 & \textbf{53.13} & 53.14 & \textbf{53.23} \\
 & box-r2 & 40.04 & \textbf{41.98} & 42.40 & \textbf{47.13} & 45.76 & \textbf{48.75} \\
 & box-r4 & 37.80 & \textbf{39.62} & 36.59 & \textbf{42.56} & 37.86 & \textbf{43.24} \\
\bottomrule
\end{tabular}
\caption{Paired plug-in query mIoU (\%) on \lvisninetytwo{} (SADe unlabeled source: FSS-1000). Raw and SADe columns report means over matched episodes; the higher value in each pair is bolded.}
\label{tab:appendix-full-lvis_fss1000_sae}
\end{table}

\paragraph{Aggregates by predictor and prompt.}
When results are aggregated by predictor, SADe has a positive mean gain for all four downstream models. FSSDINO, INSID3, and SANSA improve in all 54 matched cells for each predictor, with mean SADe--Raw gains of \PluginFSSDINOMeanDelta{}, \PluginINSIDMeanDelta{}, and \PluginSANSAMeanDelta{} mIoU, respectively; for FSS-SAM3, the mean gain is \PluginFSSSAMMeanDelta{} mIoU. The first three predictors consume the cleaned mask directly, whereas FSS-SAM3 receives its derived bounding box. SADe therefore improves every matched cell under native mask interfaces and retains a positive mean gain after conversion to FSS-SAM3's box interface, without modifying downstream inference.

\begin{table}[!tb]
\centering
\small
\setlength{\tabcolsep}{1mm}
\begin{tabular}{@{}cc|rr|rr|rr@{}}
\toprule
\multirow[c]{2}{*}[-0.45ex]{Model} & \multirow[c]{2}{*}[-0.45ex]{Prompt} & \multicolumn{2}{c|}{1-shot} & \multicolumn{2}{c|}{5-shot} & \multicolumn{2}{c}{10-shot} \\
\cmidrule(lr){3-4}\cmidrule(lr){5-6}\cmidrule(l){7-8}
 & & \multicolumn{1}{c}{Raw} & \multicolumn{1}{c|}{SADe} & \multicolumn{1}{c}{Raw} & \multicolumn{1}{c|}{SADe} & \multicolumn{1}{c}{Raw} & \multicolumn{1}{c}{SADe} \\
\midrule
\multirow{3}{*}{FSSDINO} & box & 31.00 & \textbf{35.62} & 37.10 & \textbf{45.67} & 38.11 & \textbf{45.50} \\
 & box-r2 & 28.09 & \textbf{34.79} & 31.51 & \textbf{43.08} & 34.01 & \textbf{44.85} \\
 & box-r4 & 24.68 & \textbf{32.50} & 28.33 & \textbf{40.10} & 30.15 & \textbf{43.08} \\
\midrule
\multirow{3}{*}{INSID3} & box & 28.15 & \textbf{35.36} & 32.42 & \textbf{42.08} & 33.56 & \textbf{43.02} \\
 & box-r2 & 24.78 & \textbf{34.02} & 27.93 & \textbf{39.87} & 29.08 & \textbf{41.09} \\
 & box-r4 & 22.67 & \textbf{32.11} & 24.68 & \textbf{36.47} & 25.58 & \textbf{37.90} \\
\midrule
\multirow{3}{*}{SANSA} & box & 20.67 & \textbf{22.51} & 22.40 & \textbf{26.77} & 23.53 & \textbf{27.91} \\
 & box-r2 & 18.05 & \textbf{21.31} & 19.65 & \textbf{23.98} & 22.33 & \textbf{26.85} \\
 & box-r4 & 18.40 & \textbf{21.85} & 19.47 & \textbf{23.14} & 21.49 & \textbf{23.36} \\
\midrule
\multirow{3}{*}{FSS-SAM3} & box & \textbf{39.38} & 39.33 & 50.10 & \textbf{50.32} & \textbf{42.32} & 41.34 \\
 & box-r2 & 37.33 & \textbf{38.26} & 43.20 & \textbf{47.85} & 38.97 & \textbf{41.17} \\
 & box-r4 & 35.84 & \textbf{37.02} & 38.34 & \textbf{44.76} & 35.42 & \textbf{38.35} \\
\bottomrule
\end{tabular}
\caption{Paired plug-in query mIoU (\%) on \isaidfive{} (SADe unlabeled source: COCO). Raw and SADe columns report means over matched episodes; the higher value in each pair is bolded.}
\label{tab:appendix-full-isaid_coco_sae}
\end{table}

\begin{table}[!tb]
\centering
\small
\setlength{\tabcolsep}{1mm}
\begin{tabular}{@{}cc|rr|rr|rr@{}}
\toprule
\multirow[c]{2}{*}[-0.45ex]{Model} & \multirow[c]{2}{*}[-0.45ex]{Prompt} & \multicolumn{2}{c|}{1-shot} & \multicolumn{2}{c|}{5-shot} & \multicolumn{2}{c}{10-shot} \\
\cmidrule(lr){3-4}\cmidrule(lr){5-6}\cmidrule(l){7-8}
 & & \multicolumn{1}{c}{Raw} & \multicolumn{1}{c|}{SADe} & \multicolumn{1}{c}{Raw} & \multicolumn{1}{c|}{SADe} & \multicolumn{1}{c}{Raw} & \multicolumn{1}{c}{SADe} \\
\midrule
\multirow{3}{*}{FSSDINO} & box & 31.00 & \textbf{35.49} & 37.10 & \textbf{45.63} & 38.11 & \textbf{46.42} \\
 & box-r2 & 28.09 & \textbf{35.08} & 31.51 & \textbf{43.60} & 34.01 & \textbf{44.21} \\
 & box-r4 & 24.68 & \textbf{32.34} & 28.33 & \textbf{39.62} & 30.15 & \textbf{42.93} \\
\midrule
\multirow{3}{*}{INSID3} & box & 28.15 & \textbf{35.45} & 32.42 & \textbf{42.15} & 33.56 & \textbf{43.34} \\
 & box-r2 & 24.78 & \textbf{34.26} & 27.93 & \textbf{39.67} & 29.08 & \textbf{41.45} \\
 & box-r4 & 22.67 & \textbf{32.24} & 24.68 & \textbf{36.10} & 25.58 & \textbf{37.95} \\
\midrule
\multirow{3}{*}{SANSA} & box & 20.67 & \textbf{22.59} & 22.40 & \textbf{27.30} & 23.53 & \textbf{28.42} \\
 & box-r2 & 18.05 & \textbf{21.68} & 19.65 & \textbf{24.31} & 22.33 & \textbf{27.74} \\
 & box-r4 & 18.40 & \textbf{21.63} & 19.47 & \textbf{23.32} & 21.49 & \textbf{24.16} \\
\midrule
\multirow{3}{*}{FSS-SAM3} & box & \textbf{39.38} & 39.03 & 50.10 & \textbf{50.12} & \textbf{42.32} & 41.52 \\
 & box-r2 & 37.33 & \textbf{38.56} & 43.20 & \textbf{47.69} & 38.97 & \textbf{41.08} \\
 & box-r4 & 35.84 & \textbf{37.33} & 38.34 & \textbf{44.95} & 35.42 & \textbf{37.78} \\
\bottomrule
\end{tabular}
\caption{Paired plug-in query mIoU (\%) on \isaidfive{} (SADe unlabeled source: iSAID). Raw and SADe columns report means over matched episodes; the higher value in each pair is bolded.}
\label{tab:appendix-full-isaid_domain_sae}
\end{table}

\begin{table}[!tb]
\centering
\small
\setlength{\tabcolsep}{1mm}
\begin{tabular}{@{}cc|rr|rr|rr@{}}
\toprule
\multirow[c]{2}{*}[-0.45ex]{Model} & \multirow[c]{2}{*}[-0.45ex]{Prompt} & \multicolumn{2}{c|}{1-shot} & \multicolumn{2}{c|}{5-shot} & \multicolumn{2}{c}{10-shot} \\
\cmidrule(lr){3-4}\cmidrule(lr){5-6}\cmidrule(l){7-8}
 & & \multicolumn{1}{c}{Raw} & \multicolumn{1}{c|}{SADe} & \multicolumn{1}{c}{Raw} & \multicolumn{1}{c|}{SADe} & \multicolumn{1}{c}{Raw} & \multicolumn{1}{c}{SADe} \\
\midrule
\multirow{3}{*}{FSSDINO} & box & 37.11 & \textbf{49.58} & 39.63 & \textbf{54.35} & 38.58 & \textbf{58.45} \\
 & box-r2 & 27.88 & \textbf{43.79} & 27.96 & \textbf{48.71} & 26.20 & \textbf{56.40} \\
 & box-r4 & 19.48 & \textbf{37.50} & 24.44 & \textbf{47.64} & 20.63 & \textbf{48.55} \\
\midrule
\multirow{3}{*}{INSID3} & box & 18.19 & \textbf{22.77} & 21.72 & \textbf{21.74} & 17.83 & \textbf{18.95} \\
 & box-r2 & 16.88 & \textbf{23.00} & 16.72 & \textbf{21.64} & 15.58 & \textbf{19.97} \\
 & box-r4 & 15.83 & \textbf{22.32} & 17.17 & \textbf{21.68} & 14.03 & \textbf{20.02} \\
\midrule
\multirow{3}{*}{SANSA} & box & 21.69 & \textbf{33.13} & 22.78 & \textbf{34.56} & 26.99 & \textbf{40.11} \\
 & box-r2 & 18.72 & \textbf{32.41} & 19.46 & \textbf{28.87} & 15.95 & \textbf{34.29} \\
 & box-r4 & 13.91 & \textbf{24.13} & 16.41 & \textbf{25.30} & 16.17 & \textbf{34.36} \\
\midrule
\multirow{3}{*}{FSS-SAM3} & box & 7.28 & \textbf{7.33} & 14.35 & \textbf{14.78} & \textbf{2.85} & 1.05 \\
 & box-r2 & 3.56 & \textbf{10.94} & 6.61 & \textbf{18.75} & 0.00 & \textbf{1.79} \\
 & box-r4 & 0.00 & \textbf{3.67} & 2.77 & \textbf{12.22} & 0.00 & 0.00 \\
\bottomrule
\end{tabular}
\caption{Paired plug-in query mIoU (\%) on Kvasir-SEG$\rightarrow$CVC-ClinicDB using COCO as SADe's unlabeled source. Raw and SADe are means over matched episodes; bold marks the higher value in each pair.}
\label{tab:appendix-full-polyp_coco_sae}
\end{table}

\begin{table}[!tb]
\centering
\small
\setlength{\tabcolsep}{1mm}
\begin{tabular}{@{}cc|rr|rr|rr@{}}
\toprule
\multirow[c]{2}{*}[-0.45ex]{Model} & \multirow[c]{2}{*}[-0.45ex]{Prompt} & \multicolumn{2}{c|}{1-shot} & \multicolumn{2}{c|}{5-shot} & \multicolumn{2}{c}{10-shot} \\
\cmidrule(lr){3-4}\cmidrule(lr){5-6}\cmidrule(l){7-8}
 & & \multicolumn{1}{c}{Raw} & \multicolumn{1}{c|}{SADe} & \multicolumn{1}{c}{Raw} & \multicolumn{1}{c|}{SADe} & \multicolumn{1}{c}{Raw} & \multicolumn{1}{c}{SADe} \\
\midrule
\multirow{3}{*}{FSSDINO} & box & 37.11 & \textbf{49.00} & 39.63 & \textbf{50.53} & 38.58 & \textbf{59.22} \\
 & box-r2 & 27.88 & \textbf{43.43} & 27.96 & \textbf{50.31} & 26.20 & \textbf{54.10} \\
 & box-r4 & 19.48 & \textbf{36.88} & 24.44 & \textbf{46.06} & 20.63 & \textbf{48.95} \\
\midrule
\multirow{3}{*}{INSID3} & box & 18.19 & \textbf{22.64} & 21.72 & \textbf{21.74} & 17.83 & \textbf{18.72} \\
 & box-r2 & 16.88 & \textbf{23.00} & 16.72 & \textbf{21.00} & 15.58 & \textbf{19.97} \\
 & box-r4 & 15.83 & \textbf{23.49} & 17.17 & \textbf{22.14} & 14.03 & \textbf{20.02} \\
\midrule
\multirow{3}{*}{SANSA} & box & 21.69 & \textbf{31.27} & 22.78 & \textbf{35.75} & 26.99 & \textbf{39.80} \\
 & box-r2 & 18.72 & \textbf{30.49} & 19.46 & \textbf{27.66} & 15.95 & \textbf{35.19} \\
 & box-r4 & 13.91 & \textbf{28.95} & 16.41 & \textbf{28.36} & 16.17 & \textbf{34.30} \\
\midrule
\multirow{3}{*}{FSS-SAM3} & box & 7.28 & \textbf{7.37} & \textbf{14.35} & 13.00 & \textbf{2.85} & 2.82 \\
 & box-r2 & 3.56 & \textbf{9.07} & 6.61 & \textbf{19.94} & 0.00 & \textbf{1.78} \\
 & box-r4 & 0.00 & \textbf{3.67} & 2.77 & \textbf{11.73} & 0.00 & 0.00 \\
\bottomrule
\end{tabular}
\caption{Paired plug-in query mIoU (\%) on Kvasir-SEG$\rightarrow$CVC-ClinicDB using Kvasir-SEG as SADe's unlabeled source. Raw and SADe are means over matched episodes; bold marks the higher value in each pair.}
\label{tab:appendix-full-polyp_domain_sae}
\end{table}

\begin{figure*}[!t]
\centering
\textbf{(a) \pascalfive{} / COCO}\par\smallskip
\includegraphics[width=0.74\textwidth]{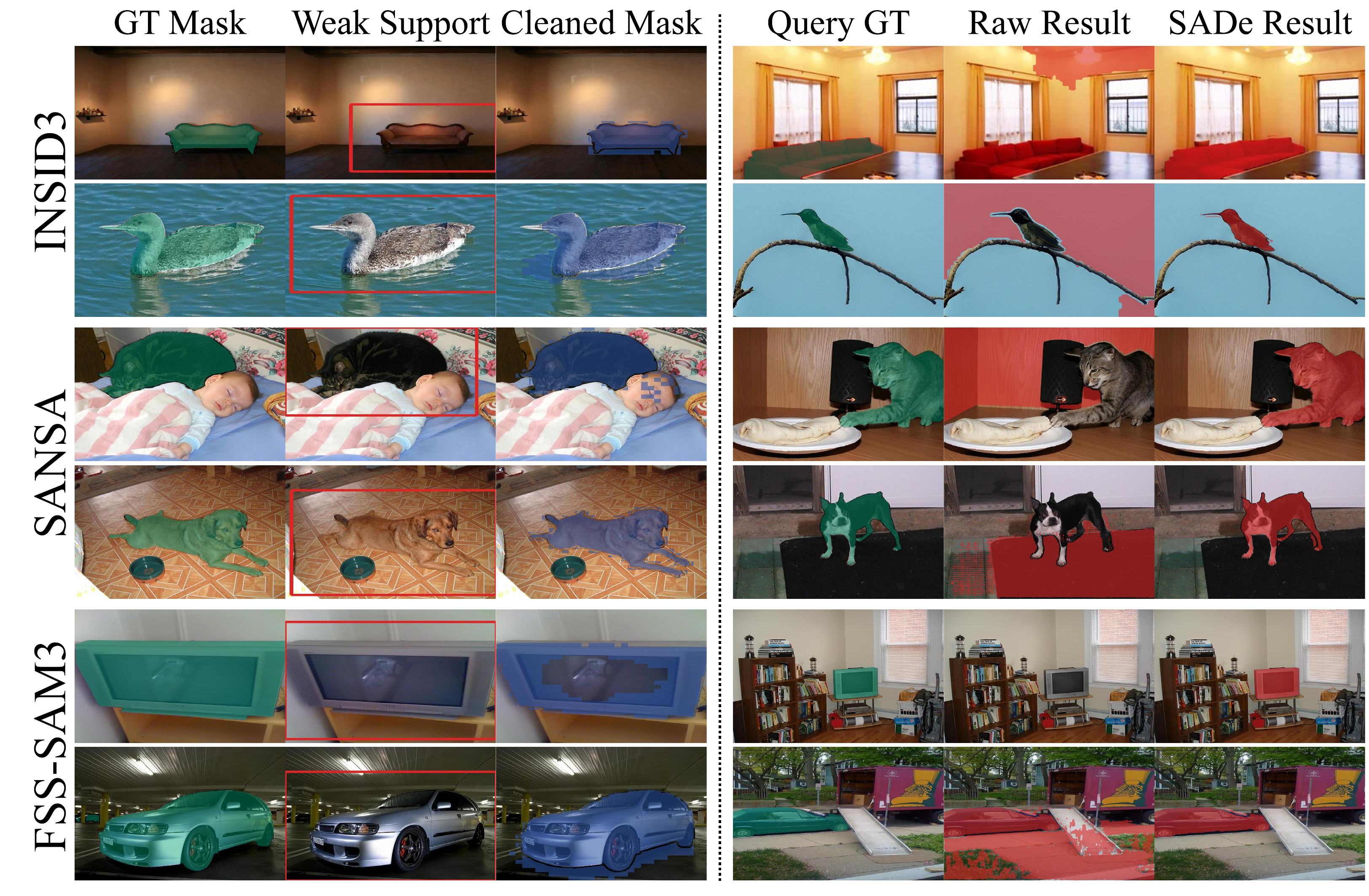}\par\medskip
\textbf{(b) \lvisninetytwo{} / FSS-1000}\par\smallskip
\includegraphics[width=0.74\textwidth]{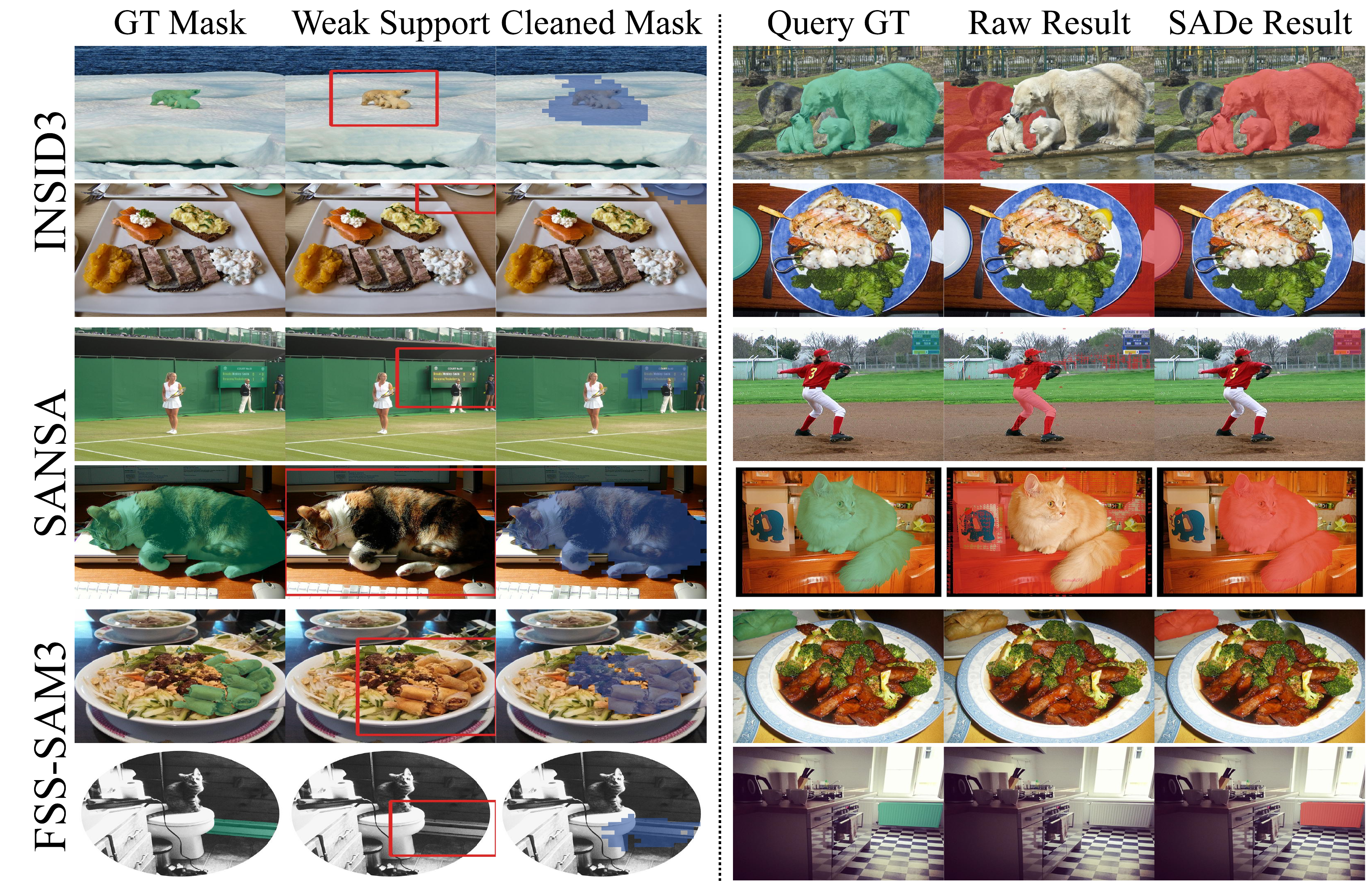}
\caption{Qualitative plug-in results on \pascalfive{} (a; unlabeled source: COCO) and \lvisninetytwo{} (b; unlabeled source: FSS-1000). Across both natural-image datasets and the displayed predictors, SADe suppresses off-target context in weak support, and its query predictions more closely follow query GT than Raw.}
\label{fig:plugin-qual-natural}
\end{figure*}

\begin{figure*}[!t]
\centering
\textbf{(a) \isaidfive{} / COCO}\par\smallskip
\includegraphics[width=0.74\textwidth]{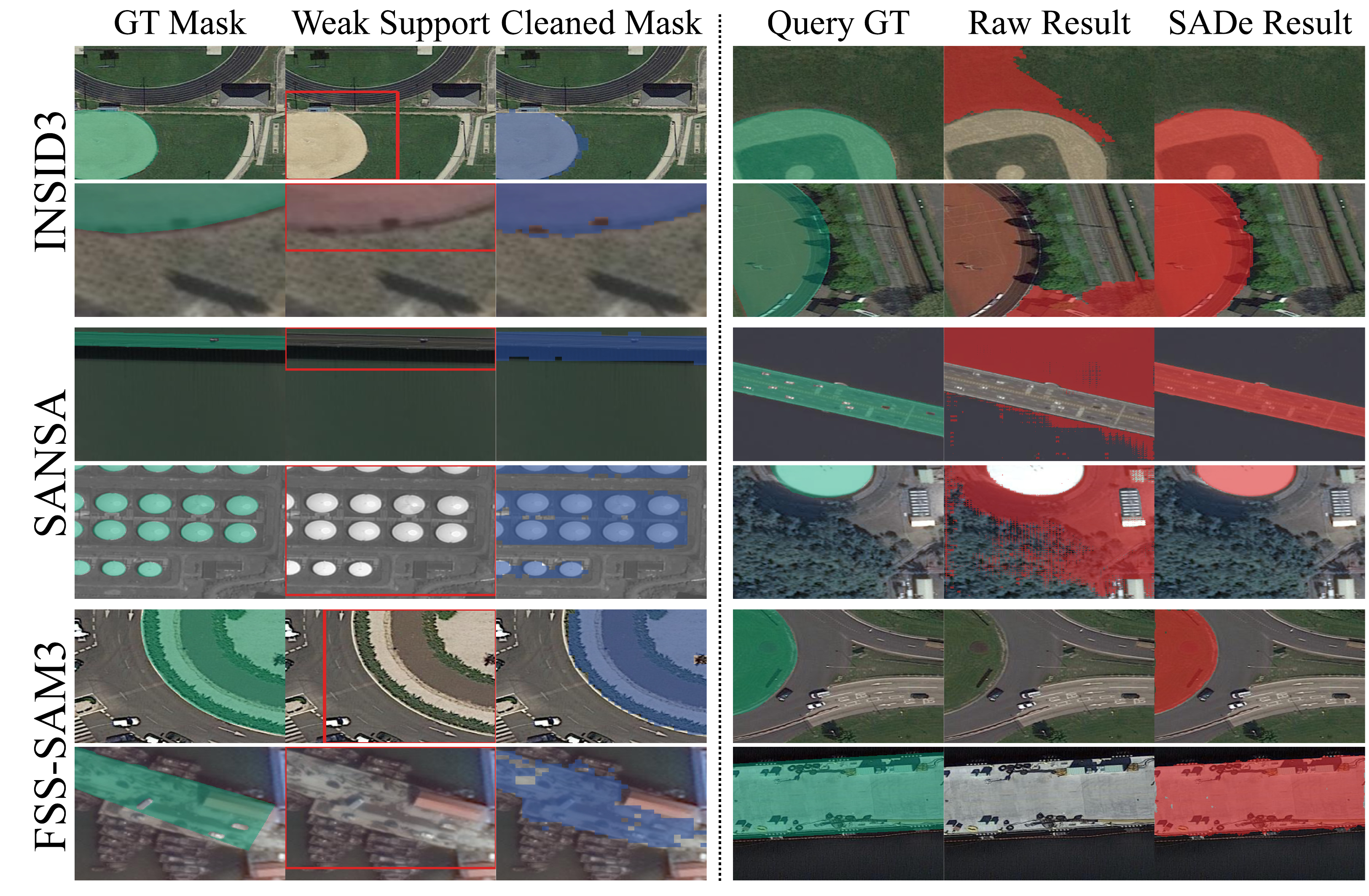}\par\medskip
\textbf{(b) \isaidfive{} / iSAID}\par\smallskip
\includegraphics[width=0.74\textwidth]{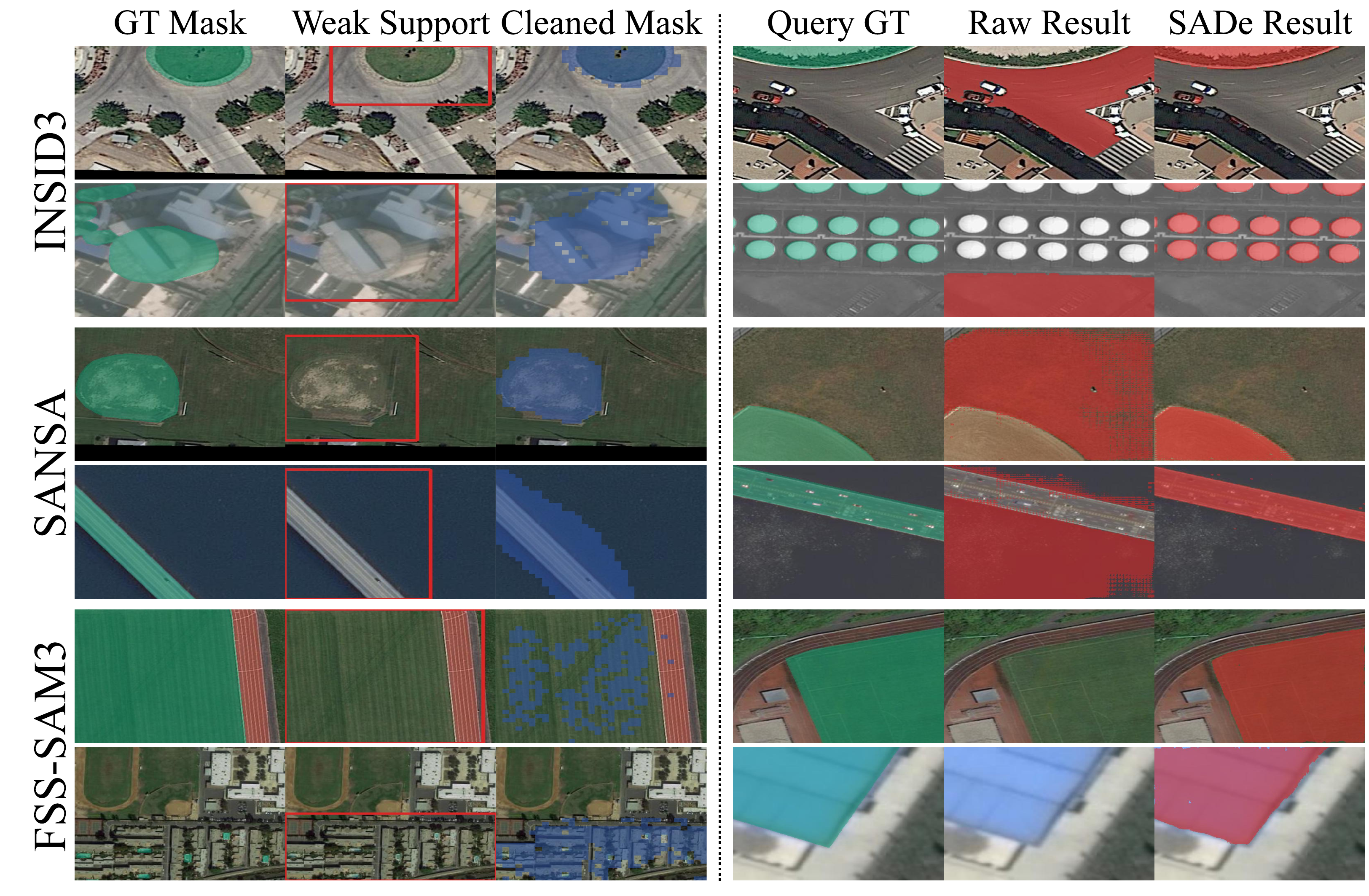}
\caption{Qualitative plug-in results on \isaidfive{} with COCO (a) or iSAID (b) as the unlabeled source. With either source, SADe removes off-target context from weak support and produces query predictions that align more closely with query GT than Raw.}
\label{fig:plugin-qual-aerial}
\end{figure*}

\begin{figure*}[!t]
\centering
\textbf{(a) Kvasir-SEG$\rightarrow$CVC-ClinicDB / COCO}\par\smallskip
\includegraphics[width=0.74\textwidth]{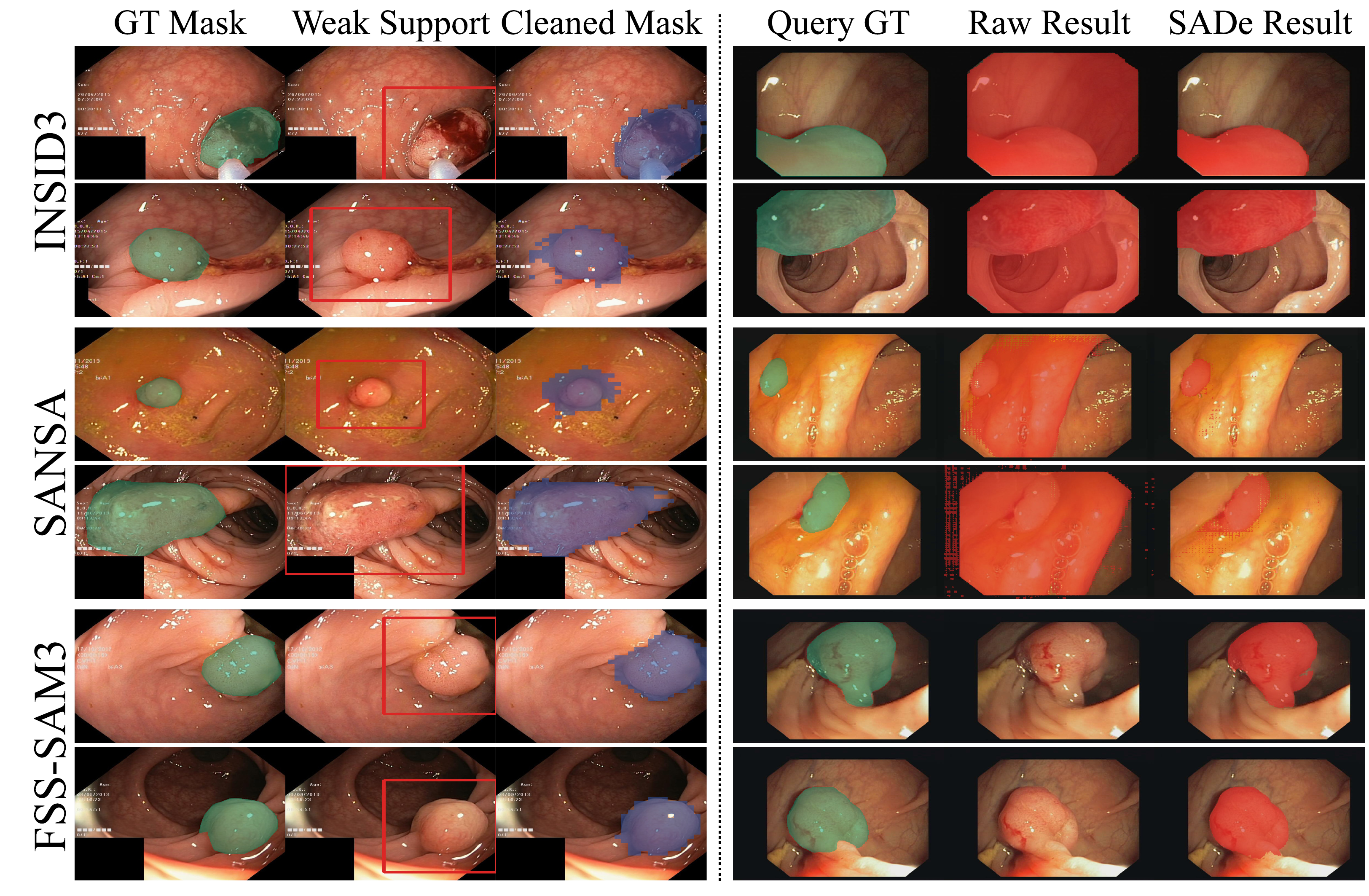}\par\medskip
\textbf{(b) Kvasir-SEG$\rightarrow$CVC-ClinicDB / Kvasir-SEG}\par\smallskip
\includegraphics[width=0.74\textwidth]{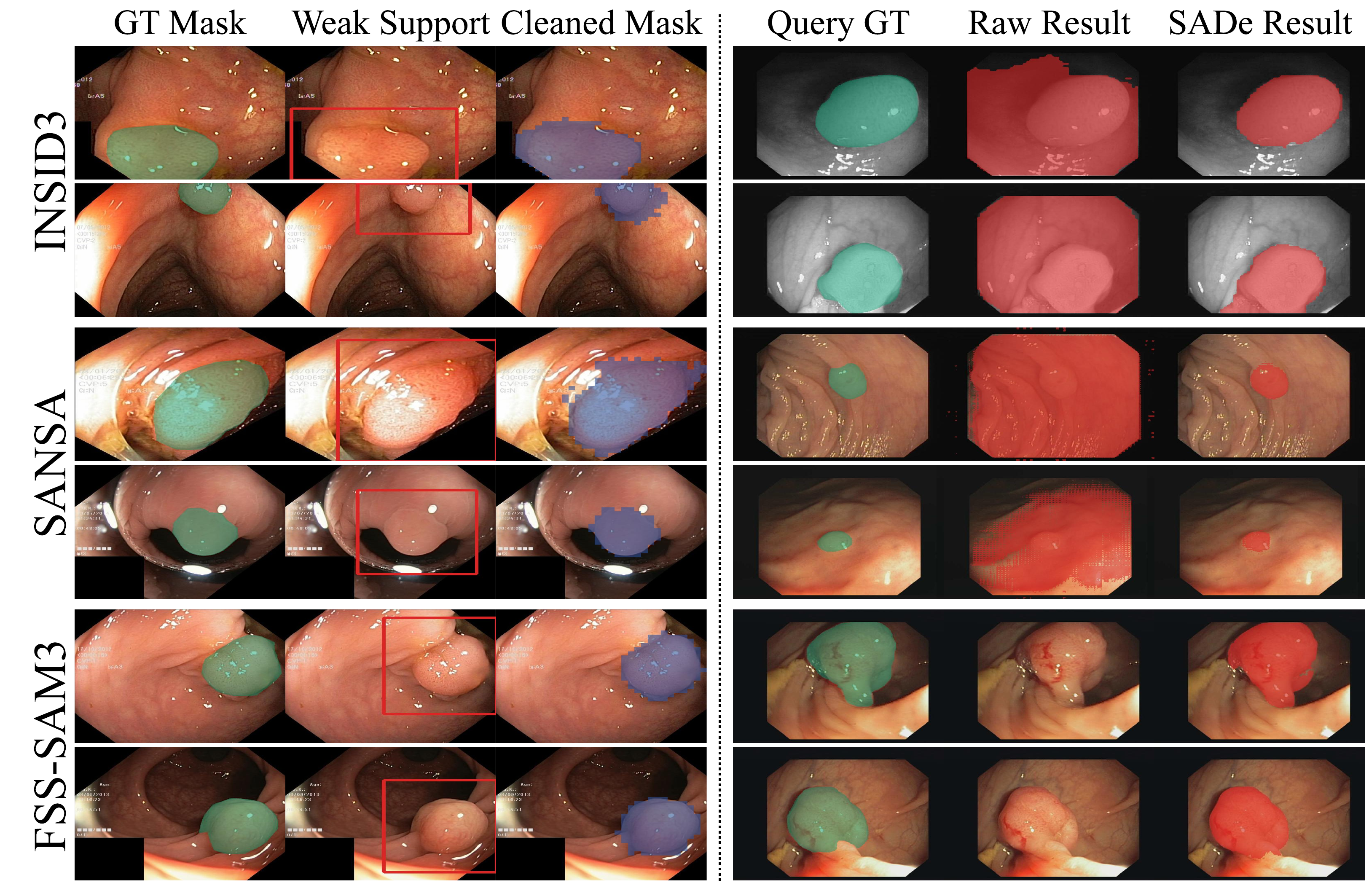}
\caption{Qualitative plug-in results from Kvasir-SEG support images to CVC-ClinicDB queries with COCO (a) or Kvasir-SEG (b) as the unlabeled source. Under both source choices, SADe concentrates the cleaned support on polyp regions and delineates query targets more accurately than Raw.}
\label{fig:plugin-qual-endoscopy}
\end{figure*}

When results are aggregated by prompt, the mean SADe--Raw gains are \PluginBoxMeanDelta{}, \PluginBoxRTwoMeanDelta{}, and \PluginBoxRFourMeanDelta{} mIoU for box, box-r2, and box-r4, respectively. All \PluginBoxRTwoCells{} box-r2 cells improve; \PluginBoxRFourWins{} of the \PluginBoxRFourCells{} box-r4 cells improve and the remaining \PluginBoxRFourTies{} tie. Expanded boxes generally contain more target-irrelevant context than tight boxes; the larger aggregate gains under box-r2 and box-r4 are therefore consistent with SADe's suppression of such context before query prediction. No box-r2 or box-r4 cell declines, so the expanded-box gains are consistent across the evaluated datasets, predictors, and shot settings rather than arising from a small subset of comparisons. Figure~\ref{fig:plugin-qual-natural} provides paired prediction examples on \pascalfive{} and \lvisninetytwo{}; Figures~\ref{fig:plugin-qual-aerial} and~\ref{fig:plugin-qual-endoscopy} show the corresponding aerial and endoscopy examples under the two source choices.

\subsection{Transfer with Generic and Domain-Related Unlabeled Sources}
\label{sec:source-transfer}

The aerial remote-sensing task compares COCO and iSAID as unlabeled image sources, while the medical-endoscopy task compares COCO and Kvasir-SEG. For each target domain, the two source configurations are compared under both the standalone and plug-in protocols, producing one source comparison per protocol. The evaluation episodes and reliability router remain fixed within each comparison; the downstream predictors are additionally fixed for the plug-in evaluation.

With COCO as the unlabeled source, SADe gains \StandaloneISaidCocoMeanDelta{} mIoU points over raw-mask ProMi in the standalone evaluation on \isaidfive{}. Its mean plug-in SADe--Raw gain is \PluginISaidCocoMeanDelta{} points, with improvements in \PluginISaidCocoWins{}/\PluginISaidCocoCells{} matched cells. The corresponding standalone and plug-in gains in medical endoscopy are \StandalonePolypCocoMeanDelta{} and \PluginPolypCocoMeanDelta{} points, with improvements in \PluginPolypCocoWins{}/\PluginPolypCocoCells{} plug-in cells. These positive gains under both protocols and in both domains show that SADe transfers directly when all source-dependent quantities used to construct dense and atom evidence are estimated solely from generic natural imagery.

Replacing COCO with domain-related unlabeled imagery increases the standalone gain from \StandaloneISaidCocoMeanDelta{} to \StandaloneISaidDomainMeanDelta{} points on \isaidfive{} and from \StandalonePolypCocoMeanDelta{} to \StandalonePolypDomainMeanDelta{} points in medical endoscopy. Meanwhile, the plug-in mean gains remain similar across source configurations: \PluginISaidCocoMeanDelta{} versus \PluginISaidDomainMeanDelta{} points on \isaidfive{}, and \PluginPolypCocoMeanDelta{} versus \PluginPolypDomainMeanDelta{} points in medical endoscopy. The corresponding improved-cell counts remain \PluginISaidDomainWins{}/\PluginISaidCocoCells{} on \isaidfive{} and change from \PluginPolypCocoWins{}/\PluginPolypCocoCells{} to \PluginPolypDomainWins{}/\PluginPolypCocoCells{} in medical endoscopy. Every configuration retains a positive gain.

Together, these results show that the frozen router is not tied to dense and atom evidence constructed from a single fixed image source: it remains effective with both generic-source evidence and evidence independently obtained from domain-related imagery.

Because the router consumes scalar atom evidence rather than atom identities, independently trained SAE dictionaries require no atom-wise alignment. The positive gains under both domain-related configurations support this alignment-free design, while the stable plug-in aggregates show that SADe's cross-predictor benefits persist across source configurations.

\end{document}